\documentclass[a4paper]{article}

\usepackage[pages=all, color=black, position={current page.south}, placement=bottom, scale=1, opacity=1, vshift=5mm]{background}
\usepackage{amsmath,amsfonts}
\usepackage{algorithm}
\usepackage{array}
\usepackage[caption=false,font=normalsize,labelfont=sf,textfont=sf]{subfig}
\usepackage{textcomp}
\usepackage{stfloats}
\usepackage{url}
\usepackage{verbatim}
\usepackage{graphicx}
\usepackage{cite}
\usepackage{hyperref}
\usepackage{overpic}
\usepackage{array,graphicx}
\usepackage{booktabs}
\usepackage{pifont}
\usepackage{libertine}
\usepackage{tabu}
\usepackage{longtable}
\usepackage{xcolor}
\usepackage{tcolorbox}
\usepackage{multicol}
\usepackage[utf8]{inputenc}
\usepackage[greek,english]{babel}
\usepackage[super]{nth}
\usepackage{subfig}
\usepackage{pgfplots}
\usepackage[super]{nth}
\usepackage{cancel}
\usepackage{arydshln}
\usepackage{longtable}

\SetBgContents{
  \parbox{\textwidth}{\tiny\tt
  © 2025 IEEE. Personal use of this material is permitted. Permission from IEEE must be obtained for all other uses, in any current or future media, including reprinting/republishing this material for advertising or promotional purposes, creating new collective works, for resale or redistribution to servers or lists, or reuse of any copyrighted component of this work in other works.
  }
}

\usepackage[margin=1in]{geometry} % full-width

\usepackage{amsmath}
\usepackage{amsthm}
\usepackage{amssymb}

\usepackage[utf8]{inputenc}
\usepackage{hyperref}
\hypersetup{
	unicode,
	pdfauthor={Stefanos Pasios, Nikos Nikolaidis},
	pdftitle={A simple article template},
	pdfsubject={A simple article template},
	pdfkeywords={article, template, simple},
	pdfproducer={LaTeX},
	pdfcreator={pdflatex}
}

\usepackage[sort&compress,numbers,square]{natbib}
\theoremstyle{plain}

\theoremstyle{definition}

\usepackage{graphicx, color}
\graphicspath{{fig/}}

\usepackage{algorithm, algpseudocode} % use algorithm and algorithmicx for typesetting algorithms
\usepackage{mathrsfs} % for \mathscr command

\usepackage{lipsum}

\title{CARLA2Real: a tool for reducing the sim2real appearance gap in CARLA simulator}
\author{Stefanos Pasios \and Nikos Nikolaidis}

\date{
	School of Informatics, Aristotle University of Thessaloniki, Thessaloniki 54124, Greece\ \texttt{\{pstefanos, nnik\}@csd.auth.gr}\\%
}

\begin{document}

        % ---------- First Page Notice ----------

            \thispagestyle{empty}
        \vspace*{\fill}
        \noindent
        \fbox{%
          \parbox{0.95\textwidth}{\footnotesize
          \textbf{This paper is accepted for publication in the IEEE Transactions on Intelligent Transportation Systems.}\\[6pt]
          S. Pasios and N. Nikolaidis, ``CARLA2Real: A Tool for Reducing the Sim2real Appearance Gap in CARLA Simulator,'' \textit{IEEE Transactions on Intelligent Transportation Systems}. \href{https://doi.org/10.1109/TITS.2025.3597010}{https://doi.org/10.1109/TITS.2025.3597010}.\\[6pt]
          © 2025 IEEE. Personal use of this material is permitted. Permission from IEEE must be obtained for all other uses, in any current or future media, including reprinting/republishing this material for advertising or promotional purposes, creating new collective works, for resale or redistribution to servers or lists, or reuse of any copyrighted component of this work in other works.
          }
        }
        \vspace*{\fill}
        \clearpage

    % ---------------------------------------

	\maketitle
	
	\begin{abstract}
		Simulators are indispensable for research in autonomous systems such as self-driving cars, autonomous robots, and drones. Despite significant progress in various simulation aspects, such as graphical realism, an evident gap persists between the virtual and real-world environments. Since the ultimate goal is to deploy the autonomous systems in the real world, reducing the sim2real gap is of utmost importance. In this paper, we employ a state-of-the-art approach to enhance the photorealism of simulated data, aligning them with the visual characteristics of real-world datasets. Based on this, we developed CARLA2Real, an easy-to-use, publicly available tool (plug-in)  for the widely used and open-source CARLA simulator. This tool enhances the output of  CARLA  in near real-time, achieving a frame rate of 13 FPS, translating it to the visual style and realism of real-world datasets such as Cityscapes, KITTI, and Mapillary Vistas. By employing the proposed tool, we generated synthetic datasets from both the simulator and the enhancement model outputs, including their corresponding ground truth annotations for tasks related to autonomous driving. Then, we performed a number of experiments to evaluate the impact of the proposed approach on feature extraction and semantic segmentation methods when trained on the enhanced synthetic data. The results demonstrate that the sim2real appearance gap is significant and can indeed be reduced by the introduced approach. Comparisons with a state-of-the-art image-to-image translation approach are also provided. The tool, pre-trained models, and associated data for this work are available for download at: \href{https://github.com/stefanos50/CARLA2Real}{https://github.com/stefanos50/CARLA2Real}.
		
		\noindent\textbf{Keywords:} Sim2real gap, CARLA simulator, Computer vision, Image processing, Autonomous driving
	\end{abstract}
	
\section{Introduction}
Simulators are fundamental to autonomous driving research as well as other related scientific domains, such as autonomous robots, since they provide a range of advantages during the development and deployment phases of such systems. Indeed, they provide a cost-effective and safe approach for the rapid prototyping of systems and algorithms. Moreover, they reduce the time required for physical testing, thus reducing the fine-tuning procedure and potentially increasing the accuracy of the final result. Additionally, real-world data are often limited as a consequence of the sparsity of some events in the real world and other time and financial constraints. The use of simulators can contribute to the generation of large and diverse amounts of synthetic data, which play a key role in achieving robustness.

While significant progress has been made in the technology and visuals of the game engines upon which these simulators are based, there is still a large gap, often referred to as the sim2real gap \cite{mahajan2024quantifyingsim2realgapgps, 10086509, yu2024naturallanguagehelpbridge, jaunet2021sim2realvizvisualizingsim2realgap, 10819363}, between the realism provided by the simulation environments and the real world. This gap has several characteristics, while NVIDIA\footnote{https://developer.nvidia.com/blog/closing-the-sim2real-gap-with-nvidia-isaac-sim-and-nvidia-isaac-replicator} divides it into two categories: the content gap \cite{metasim} and the appearance gap \cite{bujwid2018gantruthunpairedimagetoimage, 10086509, 9915679}. The first refers to the difference that may exist between the distribution of objects in the virtual and real world, such as position, frequency of appearance, scale, geometry, and content. These differences may result from variations in the geographical area or the general domain. On the other hand, the appearance gap describes the pixel-wise difference between the simulator images generated by the visual sensors and the real-world ones. Many simulators, especially open-source ones, primarily rely on low-cost, lower-quality assets. Moreover, they are often built on older or open-source engines that do not support or implement newer technologies such as real-time ray tracing. Thus, the appearance gap remains substantial, and this is clearly reflected in the datasets produced by these simulators.

Considering that the ultimate goal is to employ visual data extracted from the simulators to train and evaluate methods that are to be deployed in the real world, much research has been conducted to translate synthetic simulator images so as to exhibit real-world characteristics and appearance by targeting existing real-world datasets \cite{liao2022kitti360, cordts2016cityscapes, vistas, m2cs}. Such research often involves deep learning-related approaches based on architectures such as Generative Adversarial Networks (GANs) \cite{goodfellow2014generative}. Since most of these approaches \cite{pix2pix, park2019semantic, zhu2020unpaired, zhang2023adding, Sushko2022} focus primarily on the general structure of the objects, their translation results are frequently characterized by visual artifacts \cite{Richter_2021, zheng2022vsait, 10819363}. This is a limiting factor when the objective is to utilize the synthetic data to achieve robustness \cite{bujwid2018gantruthunpairedimagetoimage}.

\begin{figure*}[t]
  \centering
  \includegraphics[width=1\textwidth]{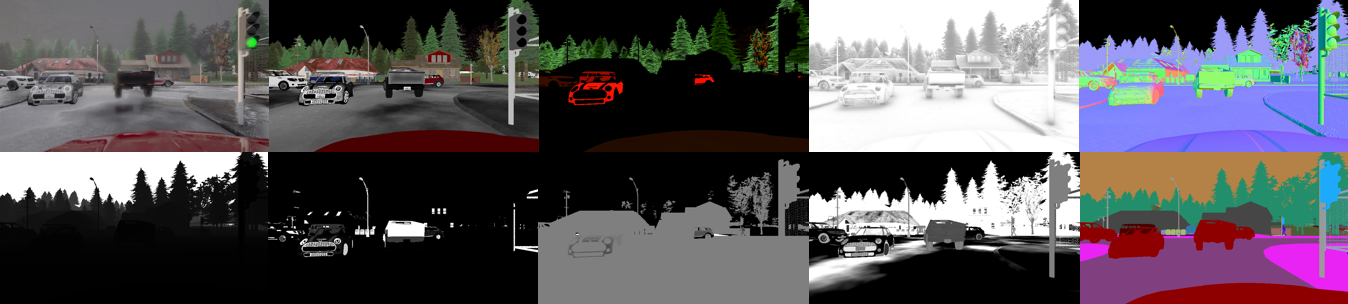}
  \caption{G-Buffers generated from the Unreal Engine 4 deferred rendering pipeline. \textbf{\nth{1} Row:} SceneColor, Albedo, Subsurface (G-Buffer D), SSAO, Normals. \textbf{\nth{2} Row:} Depth, Metallic, Specular, Roughness, Custom Stencil (Cityscapes Scheme).}
  \label{fig:carla_g_buffers}
\end{figure*}

In this paper, we focus on employing a state-of-the-art (SotA) image-to-image translation approach \cite{Richter_2021} that processes intermediate information from the rendering pipeline of the simulator engine. Such information is stored in buffers commonly referred to as Geometry Buffers or G-Buffers. Our ultimate goal is to generate robust visual data that faithfully follows the complexity of the real world and thus reduces the sim2real appearance gap. These buffers (Fig. \ref{fig:carla_g_buffers}) contain rich information about the materials, the lighting, and the geometry of the virtual objects, thus reducing visual artifacts and increasing the translation robustness. In detail, the contribution of the G-Buffers has been proven essential in neural rendering architectures such as NVIDIA's ray reconstruction technology\footnote{https://www.nvidia.com/en-us/geforce/news/gfecnt/20238/nvidia-dlss-3-5-ray-reconstruction/}. Moreover, they have been successfully applied in image-to-image translation tasks \cite{9893673, Richter_2021, 3555442} that target the enhancement of synthetic images with the characteristics of real-world datasets such as Cityscapes \cite{cordts2016cityscapes}, KITTI \cite{liao2022kitti360}, and Mapillary Vistas \cite{vistas}. However, the implementation of such an approach can be challenging, as it requires various inputs that are integrated deep into the engine. For this reason, we chose the CARLA simulator \cite{dosovitskiy2017carla}, which is open-source and provides an API that transfers the intermediate G-Buffers from the engine to the model environment. This enabled the generation of a synthetic dataset that was employed for the training of a novel visual enhancement module for CARLA. The developed module was integrated into the simulator for (near) real-time inference,  which also posed challenges due to the limitations of CARLA and the scale of the model. To achieve near-real-time performance, we introduce in our integration a multithreaded synchronization pipeline. A modified set of parameters and G-Buffers that enable higher throughput while preserving the overall output quality was also introduced. To demonstrate the impact of the introduced approach, we compared the similarity of features extracted from real-world data with features extracted from widely used architectures \cite{simonyan2015deep, he2015deep, xception} when fed with synthetic data from CARLA prior to and after the enhancement. Furthermore, we investigated how the enhancement affects the accuracy of a semantic segmentation model (a task that is closely related to autonomous driving \cite{chen2017rethinking, 10210278, XIONG2024167}) when trained on the enhanced dataset as opposed to the original CARLA data. We additionally compared the performance of the proposed approach in both feature analysis and semantic segmentation against that of a SotA image-to-image translation method \cite{zheng2022vsait} that does not employ G-Buffers. Our contributions can be summarized as follows:

\begin{enumerate}
    \item We introduce an open-source tool designed for the CARLA simulator, which aims to facilitate research in autonomous driving and reduce the sim2real appearance gap associated with the RGB camera sensor.
    \item We propose a publicly available dataset that contains additional information (G-Buffers) generated through the game engine and can be employed for training robust image-to-image translation methods that are capable of enhancing the photorealism of synthetic data.
    \item Through extensive experiments, we show that the proposed tool significantly reduces the sim2real appearance gap on feature extraction and semantic segmentation methods, and compare it against a SotA image-to-image translation approach that does not utilize additional G-Buffers. 
\end{enumerate}

\section{Related Work}\label{sec2}

Few studies have investigated whether visual simulation data can be enhanced by GAN-based \cite{NIPS2014_f033ed80, park2020contrastive, tang2021attentiongan, zhu2020unpaired, han2021dual, liu2018unsupervisedimagetoimagetranslationnetworks, Richter_2021, pix2pix} methods in order to reduce the sim2real appearance gap and thus be utilized in training algorithms for autonomous driving in real-world scenarios. Particularly, there is a significant focus on computer vision-related tasks employed in autonomous vehicles, including semantic segmentation \cite{chen2018encoderdecoder, hou2019learning}, feature extraction \cite{kingma2022autoencoding, simonyan2015deep, he2015deep, xception}, and deep reinforcement learning \cite{lillicrap2019continuous} using visual input from the ego vehicle camera.

Bujwid et al. \cite{bujwid2018gantruthunpairedimagetoimage} proposed GANtruth, an image-to-image translation method aimed at preserving the semantic content of the original synthetic images in order to maintain consistency with the ground truth (GT) labels. The authors compared their method against two image-to-image translation approaches \cite{NIPS2014_f033ed80, liu2018unsupervisedimagetoimagetranslationnetworks} with the goal of translating synthetic images from the SYNTHIA \cite{7780721} dataset towards Cityscapes. It was observed that, when training a DeeplabV3+ \cite{chen2018encoderdecoder} semantic segmentation model on the translated data and testing on the Cityscapes validation set, the proposed method could marginally surpass the model trained on the original synthetic data. Also, the baseline image-to-image translation methods illustrated substandard performance due to the significant visual artifacts they introduce. Tsogkas et al. \cite{gta_deeplabv3} performed a similar study by employing more recent image-to-image translation methods \cite{park2020contrastive, tang2021attentiongan, zhu2020unpaired, han2021dual} to translate frames from the Grand Theft Auto V (GTAV) game \cite{Richter_2016_ECCV} so as to match a real urban environment \cite{cordts2016cityscapes}. The results indicated that the accuracy of a semantic segmentation network such as DeeplabV3+ is slightly higher when training on the translated synthetic frames and testing on real data, compared to training on the initial synthetic data and testing on the real world. The limited improvement is attributable to the visual errors introduced by the specific translation models \cite{gta_deeplabv3}. If these occur frequently and to a large extent, then the negative effect on accuracy is intensified. That's because the GT labels will correspond to different objects or artifacts (e.g., misplaced trees in the sky due to distributional differences between the synthetic dataset and the real-world dataset \cite{Richter_2021}) incorrectly generated from the translation method.

Pahk et al. \cite{10086509} employed Dual Contrastive Learning Generative Adversarial Network (DCLGAN) \cite{han2021dual} for a Lane Keeping Assist System in CARLA. In detail, DCLGAN was utilized to translate frames that solely contained the road of the CARLA simulation environment to mimic the characteristics of real images captured from the Korea Intelligent Automotive Parts Promotion Institute (KIAPI). The translated frames achieved a lower Frechet Inception Distance (FID) \cite{fid} and improved the lane segmentation accuracy of ENet-SAD \cite{hou2019learning} when applied to the real-world images. Moreover, the ability of the system for lane restoration was proved to improve when employing the translated images. Ram et al. \cite{ddpg-sim2real} evaluated the Pix2Pix \cite{pix2pix} method in conjunction with a Deep Deterministic Policy Gradient Reinforcement Learning Algorithm \cite{lillicrap2019continuous} for autonomous driving within the CARLA simulator \cite{dosovitskiy2017carla}. The model was trained to synthesize images resembling Cityscapes based on semantic GT label maps extracted from the simulator. Similar to  \cite{gta_deeplabv3, bujwid2018gantruthunpairedimagetoimage}, it was observed that there is almost no improvement compared to using the raw synthetic data from the simulator. Despite maintaining the overall frame structure, the quality of the synthesized images remained subpar, characterized by the presence of visual artifacts.

 Subsequently, more robust methods were employed for conducting such comparisons \cite{Richter_2021}. In \cite{9915679}, a feature analysis was conducted on GTAV  frames translated so as to target the characteristics of Cityscapes and Mapillary Vistas, along with synthetic datasets originating from CARLA \cite{dosovitskiy2017carla, Alberti_2020} and Virtual KITTI \cite{gaidon2016virtual}, comparing their similarity with Cityscapes. To extract features, conventional pre-trained classification models such as VGG-19 \cite{simonyan2015deep} and RESNET-152 \cite{he2015deep} were employed and fine-tuned for the task of classifying the origin dataset of each input frame. The feature similarity was evaluated as the distance between the centroids of the clusters of each dataset and the centroid of the Cityscapes frames after applying Principal Component Analysis (PCA) to reduce the dimensions to two. It was observed that, indeed, the features of the enhanced synthetic datasets are closer to the Cityscapes features compared to those of raw synthetic data. However, applying PCA to high-dimensional data can lead to information loss. The comparison in 2D space is incomplete and cannot convey comprehensive information about the actual distance and gap that exist in the original high-dimensional feature space. Furthermore, the absence of corresponding translated datasets and the comparison of synthetic datasets that exhibit different levels of visual realism due to the game engine and assets used (e.g., textures, lighting, or 3D models) can possibly introduce potential bias in the comparison. Indeed, GTAV \cite{Richter_2016_ECCV} is a commercial game whose quality is higher than that of CARLA, an open-source simulator mainly built on low-quality assets. A part of the gap reduction may result from that quality difference and not the actual image-to-image translation method.

\section{CARLA Enhancement}\label{sec3}

In this section, we discuss the methodologies and tools involved in our research, which focuses on the integration of a robust image-to-image translation method towards reducing, in real time,  the sim2real appearance gap that exists in CARLA. We begin by briefly describing the Enhancing Photorealism Enhancement \cite{Richter_2021} and VSAIT \cite{zheng2022vsait} approaches. Then, we delve into the real-world datasets that were employed in our work, along with our CARLA synthetic dataset extraction and training strategy. Finally, we introduce the CARLA2Real tool along with its limitations and illustrate the translation results that can be achieved by utilizing the tool.

\subsection{Enhancing Photorealism Enhancement (EPE)}\label{subsec2}

Enhancing Photorealism Enhancement (EPE) \cite{Richter_2021} is a robust photorealism enhancement approach proposed by Richter et al. that translates synthetic data generated from a virtual environment towards real-world urban datasets such as Cityscapes, KITTI, and Mapillary Vistas. The introduced approach, in contrast to the previous works, employs the intermediate G-Buffers that are generated from the rendering pipeline in each frame. These buffers include useful information about the geometry, materials, and lighting of the scene, which are important to reduce the risk of visual artifacts. 

The training of the enhancement network is divided into two main objectives. The Learned Perceptual Image Patch Similarity (LPIPS) loss, which calculates the similarity between images and penalizes structural dissimilarities, and the perceptual discriminator that provides a realism score. Specifically, the perceptual discriminator employs semantic information extracted from the real and synthetic datasets from a robust semantic segmentation network and extracts features from a VGG-16 \cite{simonyan2015deep} architecture for the real and enhanced images. To generate compatible semantic segmentation annotations across domains, EPE employs the Multi-Domain Segmentation (MSeg) network \cite{mseg1} as a robust pre-trained model applied prior to training. MSeg is used to produce approximately consistent label maps for both the real and synthetic datasets. Thus, it enables the extraction of semantic information for real images that lack GT annotations, which are often difficult or costly to obtain.

The G-Buffers, along with the corresponding GT label maps of the synthetic dataset, are processed by a G-Buffer encoder network. The GT label maps are preprocessed into a set of one-hot encoded masks for a particular class or a group of classes. The G-Buffers can potentially not contain any valuable information for a particular class. This is the case with the sky, where no geometry exists. Thus, the G-Buffer encoder consists of multiple distinct streams that process the G-Buffers and can treat each individual object (class) differently. To reduce the number of streams, similar objects in the semantic map, such as vehicles (Car, Truck, Bus, Train, Motorcycle, and Bicycle), are grouped into a single one-hot encoding mask. The resulting tensor is further processed by residual blocks that extract feature tensors at multiple scales before each downsampling. The feature tensors are then processed by the enhancement network via the Rendering-Aware Denormalization (RAD)  modules.

The distribution of visible objects can differ between the real and synthetic datasets, and this may lead to visual artifacts, since the discriminator can easily distinguish whether a frame is real or fake based on that difference. Thus, a patch-matching approach is pursued. For both the synthetic and real images, patches are generated and processed via a VGG-16 network for feature extraction on those crops. For features extracted from each crop in the synthetic dataset, the k-nearest neighbors algorithm is applied by employing the Facebook AI Similarity Search (FAISS) library \cite{faiss}. The neighbors are finally filtered by a distance threshold based on the calculated $L_{2}$ distance during the FAISS similarity search.

\subsection{VSAIT}

 Theiss et al. \cite{zheng2022vsait} introduced an unpaired image-to-image translation method, VSAIT, aiming to mitigate semantic flipping, which includes the alteration of the content of the source images, such as transforming the sky into trees. To achieve this, VSAIT employs Vector Symbolic Architectures (VSA) and introduces VSA-based constraints into adversarial learning, enabling the model to learn a mapping that can invert the translation process. The method was applied to synthetic data extracted from GTAV \cite{Richter_2016_ECCV} and translated towards the characteristics of Cityscapes. To evaluate the contribution of the proposed method, a Cityscapes-pretrained DeepLabV3 \cite{chen2017rethinking} model was applied on the enhanced data along with the respective images produced by various SotA unpaired image-to-image translation methods. Due to the lack of publicly available datasets that incorporate additional information extracted from the game engines, the authors compared the proposed method against EPE both qualitatively and quantitatively, using data reported in \cite{Richter_2021}, without including the approach in the semantic segmentation experiments. Along these lines, it was observed that EPE exhibited issues such as removing palm trees and adding unnatural lighting on cars. Additionally, the Kernel Inception Distance (KID) metric \cite{bińkowski2018demystifying}, calculated between the original GTAV and the respective translated images, was proved to be slightly lower compared to the one reported in \cite{Richter_2021}.

\subsection{Real-World Datasets} \label{real-world-datasets}

Training the EPE model effectively requires real-world datasets that share the same layout and distribution of objects as CARLA, which simulates urban environments. Richter et al. \cite{Richter_2021} trained the model to enhance frames extracted from a virtual game, GTAV \cite{Richter_2016_ECCV}, which also takes place in a fictional urban environment based on the city of Los Angeles. G-Buffers were directly extracted from the GPU, where they are natively generated by the game engine. The authors employed three publicly available urban environment datasets: Cityscapes, KITTI, and Mapillary Vistas. These datasets are also suitable for the translation of the CARLA simulator output and thus were also used in our work.

Cityscapes \cite{cordts2016cityscapes} is a publicly available dataset typically employed for semantic segmentation-related tasks, as it provides pixel-level GT annotations for $30$ distinct semantic classes. The dataset contains $5,000$ images and is extended to $20,000$ images with coarse (less detailed) annotations. It is geographically restricted in Germany, and the images were captured in $50$ cities in good weather conditions in daylight. A characteristic of the Cityscapes dataset is that the hood of the ego vehicle is visible throughout the entire dataset.

KITTI \cite{liao2022kitti360} is another dataset widely used in autonomous driving research. It is even more geographically restricted compared to Cityscapes as it was captured entirely in the city of Karlsruhe, Germany. The dataset provides $15,000$ images along with autonomous driving-related system information such as laser scans, high-precision GPS measurements, and IMU accelerations.

In contrast to the Cityscapes and KITTI datasets, the Mapillary Vistas \cite{vistas} dataset extends its diversity to cities across six continents. It contains $25,000$ high-resolution images, while it is accompanied by pixel-wise GT labels for $124$ different semantic and $100$ instance categories.

\subsection{Synthetic Dataset} \label{synthetic_dataset}

Both real and synthetic datasets are required for training the enhancement method. Since CARLA provides an accessible way to retrieve G-buffers information, a publicly available synthetic dataset consisting of 20.014 images was generated (Table \ref{dataset_details}) with the goal of training the photorealism enhancement model.

\begin{table}[!htb]
  \footnotesize
    \caption{\label{dataset_details} Details about the characteristics of the generated synthetic dataset.}
  \centering
  \begin{tabular}{@{}ll@{}}
    \toprule
    \textbf{Dataset Characteristics} & \textbf{Details} \\
    \midrule
    Available Data & RGB Frames (PNG), G-Buffers (NPZ/PNG), \\ & Ground Truth Labels (PNG).\\
    Camera Resolution & $960x540$ \\
    Dataset Size & $119$ GB (unpreprocessed) \\
    Number of Frames & $20,014$ ($15,011$ + $5,003$ frames\\
    & including nighttime) \\
    Weather Presets & ClearNoon, CloudyNoon, WetNoon,\\ & ClearSunset, MidRainNoon, \\ & HardRainNoon,  SoftRainNoon, \\ & WetCloudyNoon, CloudySunset,\\ &
    WetSunset, Fog, WetCloudySunset,\\ & MidRainSunset, WetCloudySunset, \\ & MidRainSunset, HardRainSunset, \\ & SoftRainSunset \\
    CARLA Towns & Town10HD, Town01, Town02, Town03, \\ & Town04, Town05 \\
    Time of Day & Noon, Sunset, Night, Default \\
    Perspectives & Front View Camera, Hood Camera \\
    Traffic & All the available CARLA vehicles and \\ &  pedestrians assets from the catalogues. \\
    \bottomrule
  \end{tabular}
\end{table}

The synthetic dataset includes RGB frames, the respective intermediate G-Buffers, and the GT label maps. These data were exported from the CARLA simulator in synchronous mode (Section \ref{syn_async_section}) through a method that ensured data synchronization through frame identifiers. This approach was adopted so as to avoid any accuracy decrease during the training phase due to any data dissimilarity that can occur in the asynchronous mode. The frames were extracted by keeping one out of $20$ frames from the camera feed of a moving vehicle to avoid having very similar images. We targeted a variety of CARLA towns, weather presets, vehicles, camera perspectives (including the ego vehicle hood), and other aspects of the environment. These were randomized in each frame or were manually controlled to preserve a balanced distribution of the dataset characteristics (e.g., CARLA towns and perspectives), considering that deep learning models are sensitive to data imbalance. 

In order to increase the applicability of the dataset, it was extracted in an unpreprocessed state. This enables its utilization not only in semantic segmentation \cite{chen2017rethinking} and G-Buffer estimation (e.g., depth \cite{ranftl2020robust} and surface normal \cite{bansal2016pixelnet} estimation) tasks but also in a variety of other image-to-image translation approaches that solely require the GT labels instead of the intermediate G-Buffers \cite{pix2pix}. Subsequently, the G-Buffers were stacked in the channels of a single tensor. The semantic segmentation GT labels were grouped in one-hot encoded masks of just 12 distinct channels from the total of 29 available semantic classes\footnote{The semantic classes and their corresponding IDs are available at: \url{https://carla.readthedocs.io/en/latest/ref_sensors/\#semantic-segmentation-camera}} of CARLA. Considering that Playing for Data \cite{Richter_2016_ECCV} and CARLA adhere to the Cityscapes annotation scheme for that particular procedure, we followed the same grouping method as suggested in \cite{Richter_2021}. In addition, all frames of the synthetic but also the real dataset were passed through the robust MSeg \cite{mseg1} semantic segmentation network. This was done in order to obtain the robust GT label maps that are important to specialize the discriminator on individual classes when calculating the realism score.

Since the photorealism enhancement model was designed to be applied to images with a resolution of approximately 960x540, we used the same resolution when extracting the CARLA dataset. In the case of the real datasets, where image resolutions vary depending on the dataset (Cityscapes and Mapillary Vistas, see subsection \ref{real-world-datasets}), all frames were uniformly resized to 960x540. However, for KITTI, where frames are in wide format, to maintain the aspect ratio and avoid object deformation or loss of information, images were kept at their original resolution of 1242x375.

\subsection{Training and EPE Modifications}

After collecting the required real-world (Section \ref{real-world-datasets}) and synthetic data (Section \ref{synthetic_dataset}), the next step involves training the photorealism enhancer. Richter et al. \cite{Richter_2021} trained the model with four total layers for both the generator and the discriminator, stating a delay of 0.5 seconds with an RTX 3090 GPU without further describing the testing conditions of this setup. Training the model with the suggested parameters and by employing the published code\footnote{https://github.com/isl-org/PhotorealismEnhancement} resulted in a delay of 1.35 seconds with an RTX 4090 GPU, an I7 13700KF CPU, and 32 GB of system memory. This obviously deviates significantly from our real-time objective. Adjusting the parameters of the generator (the component that is utilized during the model inference) to a single layer reduced the inference time to 0.3 seconds. The reduction of the generator’s parameters also led us to reduce the parameters of the discriminator to three layers so as to maintain stability during the training phase. Despite the reduction of the parameters, it was observed that the quality of the translations maintained an adequate level, demonstrating that the model could achieve a faster inference time without compromising translation effectiveness.

Furthermore, in order to minimize the risk of the model learning spurious features from noise, a selective approach was followed. In more detail, we filtered the G-Buffers so as to process, through the G-Buffer Encoder streams, only those containing relevant information for each of the 12 compressed one-hot encoded masks. For example, as already discussed by Richter et al. \cite{Richter_2021}, most of the G-Buffers do not contain any useful information about the sky. Indeed, the corresponding pixels are always zero-valued due to the fact that the sky does not contain any geometry. Additionally, in the depth buffer, pixels corresponding to that particular class always appear white due to the significant distance between the sky and the camera.

Through further experimentation, it was determined that G-Buffer D was only suitable for the Vegetation and Vehicles classes, as it provided useful information for these objects, while pixels for other classes were consistently black (zero value). For the sky class, the only available information could be obtained from the SceneColor buffer. The velocity buffer was entirely removed from consideration, as it was observed that it did not contribute to the final translation result. Since the data transfers between the TCP connection and the hardware can be computationally expensive, removing some buffers can potentially improve performance.

Training for all the models was conducted for up to $600,000$ iterations, and the training process could last up to three consecutive days. The model is designed to be trained with a batch size of one, loading all the inputs directly from the disk for each iteration due to memory constraints, thus adding an extra training performance bottleneck. It was observed that beyond the range of $360,000$ to $400,000$ iterations, the models tended to overfit, introducing artifacts in the translation results. In the case of the Cityscapes dataset, this effect was particularly noticeable in trees.

\begin{figure}[t]
  \centering
  \includegraphics[width=\textwidth]{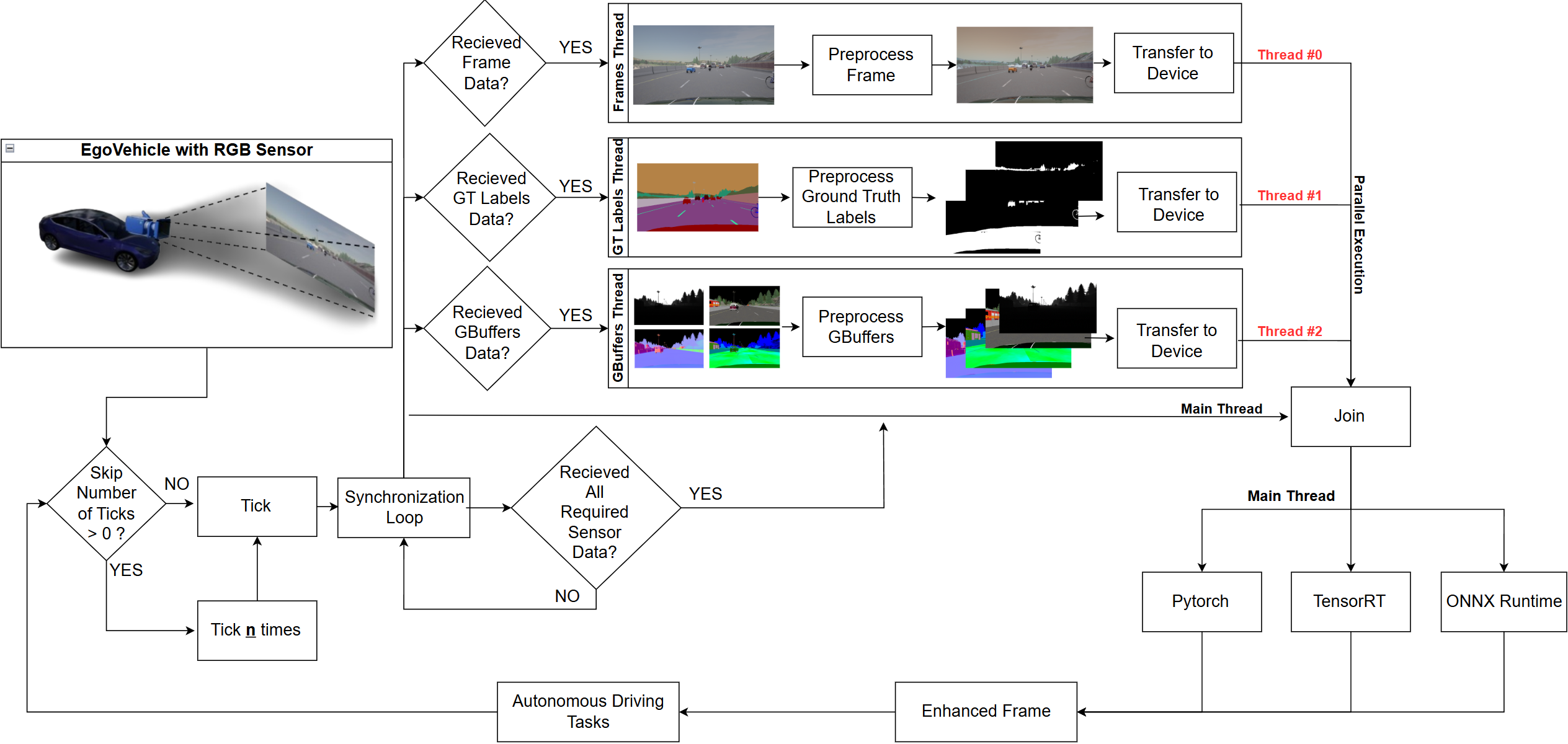}
  \caption{The EPE integration flowchart in the CARLA simulator for synchronous mode.}
  \label{fig:flowchart}
\end{figure}

\subsection{CARLA2Real}

The CARLA2Real tool was developed with the objective of providing a variety of features that enable procedures that are fundamental for autonomous driving, such as synthetic dataset generation, autonomous driving algorithms, and related tasks. This was made possible by solving numerous CARLA challenges, including the sim2real appearance gap that exists between the Unreal Engine 4 rendering pipeline and the real world. To the best of our knowledge, there is no previous work that integrates the EPE model \cite{Richter_2021} on the output of a virtual simulator or game in real-time. This can be attributed to the fact that  EPE requires numerous inputs that are integrated deep into the engine and high computational resources to run in near-interactive (non-real) time. Our implementation tried to overcome these computational limitations by focusing on performance improvements and optimizations through multithreading and other SotA tools.

\subsubsection{YAML Parameterization}
Our approach targets the creation of an accessible and easy-to-use plug-in for the CARLA simulator. Thus, all the implemented functionalities and the parameters of the CARLA simulator are controlled through a parameterizable YAML file and a set of specifically crafted scripts that are related to the supported tasks. These provide a unified configuration without requiring direct and cumbersome interaction of the user with the core components of the implementation. In more detail, we provide a set of parameters that are related to: 
\begin{itemize}
    \item various aspects of the environment (town, weather, ego vehicle, sensors, etc.)
    \item the performance of the approach (data precision type and compilers) \item autonomous driving (semantic segmentation \cite{chen2018encoderdecoder, chen2017rethinking}, object detection \cite{yolo}, reinforcement learning \cite{lillicrap2019continuous, mnih2013playing}, and other related autonomous driving models \cite{bojarski2016end}) 
    \item data generation functionalities (data format, annotations, and real-time visualization). 
\end{itemize}
Considering that one of the goals of our work was the evaluation of the sim2real appearance gap, we also provided the option to utilize the tool features outlined in this section on both the original and the enhanced simulator output. The option to completely disable the inference with the EPE model to improve the performance is also included.

\subsubsection{Synchronous \& Asynchronous Modes} \label{syn_async_section}

\begin{figure}[t]
  \centering
  \includegraphics[width=0.6\textwidth]{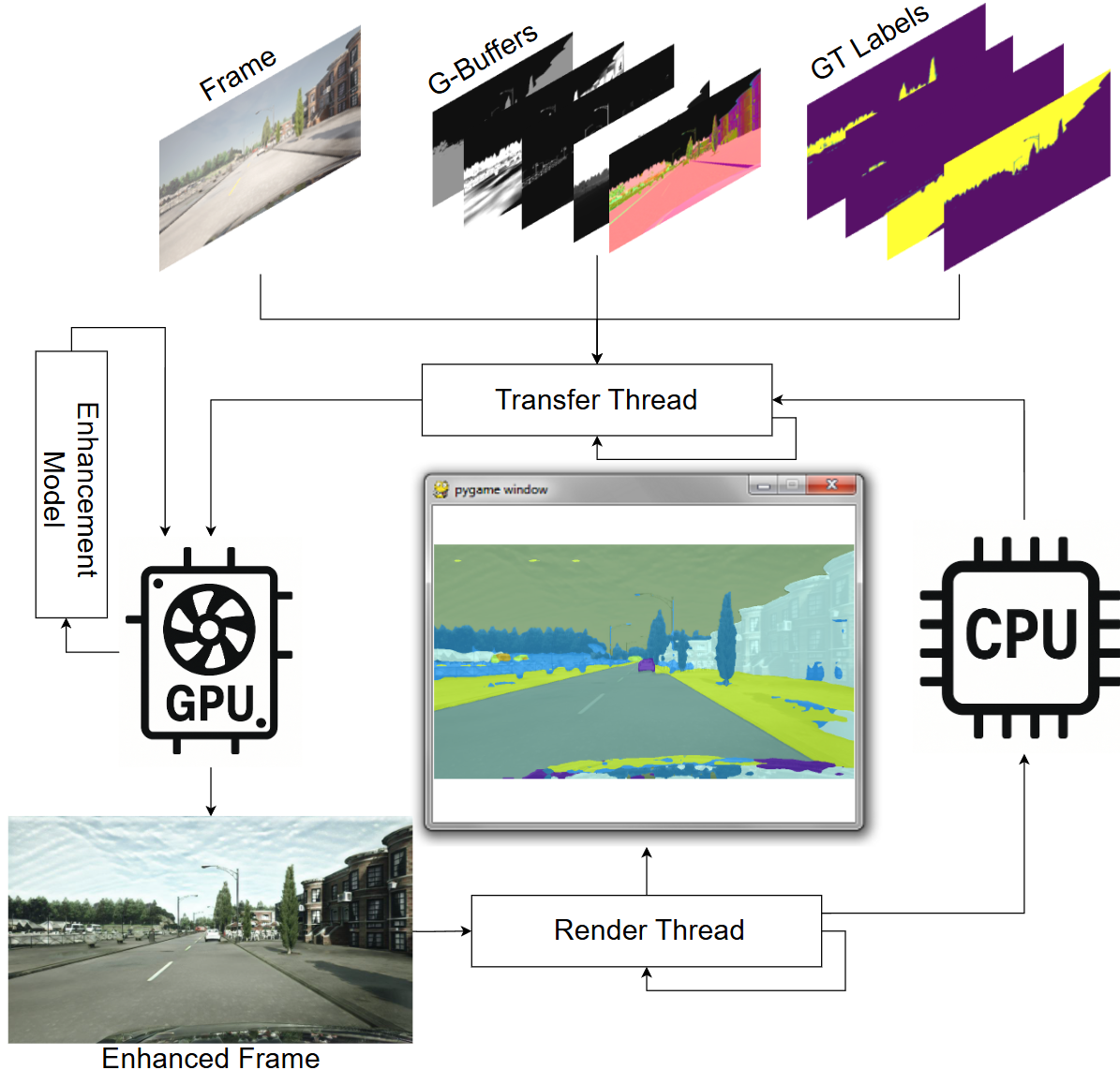}
  \caption{The EPE integration flowchart in the CARLA simulator for asynchronous mode.}
  \label{fig:flowchart2}
\end{figure}

CARLA supports the execution of the world in a synchronous and asynchronous manner. The first approach disables the multithreading on the engine side, thus reducing the performance of the simulation, yet it validates the synchronization and integrity of the data that are transferred between the server and the client. On the other hand, the asynchronous approach favors the rapid execution of the simulation. However, there is no synchronization guarantee between the data that are transferred from the engine to the Python environment, which can lead to potential issues during the training or evaluation phases of an autonomous driving algorithm. As illustrated in Fig. \ref{fig:flowchart} and Fig. \ref{fig:flowchart2}, the CARLA2Real tool supports both approaches by ensuring data integrity, particularly between the G-Buffers, which CARLA's queue-based synchronous pipeline \footnote{\url{https://carla.readthedocs.io/en/latest/adv_synchrony_timestep/}} does not currently support. To be specific, our synchronous pipeline utilizes multithreading in order to parallelize the preprocessing and data transfer steps and reduce the overall execution delay. In detail, as illustrated in Fig. \ref{fig:flowchart}, after a specified number of world ticks, the proposed pipeline waits to collect three types of data from the RGB sensor of the ego vehicle: frames, G-buffers, and GT labels. These data streams are received, processed, and transferred to the GPU asynchronously by a dedicated thread for each data type. Once all required data have been processed, they are fed to the photorealism enhancement model. The resulting enhanced frame is subsequently used for the autonomous driving task (i.e., semantic segmentation, object detection, or reinforcement learning). In contrast, in the asynchronous mode, we adhered to an asynchronous data transfer pipeline (Fig. \ref{fig:flowchart2}) where the GPU asynchronously receives any of the preprocessed data (frames, G-buffers, or GT labels) from a transfer CPU thread. Thus, during the next inference, the required inputs of the photorealism enhancement model will already be loaded into the GPU memory, reducing the delay of joining the pre-processing threads. Subsequently, the render thread is a dedicated thread that constantly receives the output of the photorealism enhancement model through a queue and renders it in a window along with any autonomous driving-related model predictions, such as a semantic segmentation overlay.

\subsubsection{Synthetic Data Generation}
Experiments of this work are focused on comparing the sim2real appearance gap of CARLA output before and after applying the EPE model in autonomous driving tasks that are mainly trained outside of the simulator. Therefore, the proposed tool implements a dataset extraction method in synchronous mode in order to extract identical frames (before and after translation) and synchronized annotations from the same simulation frame. As illustrated in Fig. \ref{fig:flowchart}, before entering the synchronization loop and receiving the required data for inference with the photorealism enhancement method, a number of frames can be skipped. This mechanism was implemented since, during data generation, the data are selected based on a timestep. Specifically, the  data that can be generated, depending on the parameterization, are:

\begin{itemize}
  \item \textbf{Semantic Segmentation:} The Custom Stencil Buffer that contains the GT label maps, as a PNG or JPG image. 
  \item \textbf{Depth:} The Depth Buffer that contains the distance between the camera and the world objects, as a PNG or JPG image. 
  \item \textbf{CARLA Frame:} The frame that was rendered from the ego vehicle camera, as a PNG or JPG image. 
  \item \textbf{Enhanced CARLA Frame:} The frame that was rendered from the ego vehicle camera after applying the EPE approach, as a PNG or JPG image. 
  \item \textbf{Object Detection Annotations:} Object detection bounding boxes supporting nine distinct classes: person, vehicle, truck, bus, motorcycle, bicycle, rider, traffic light, and traffic sign in Pascal VOC format. 
  \item \textbf{Vehicle Status:} Information about vehicle controls (steer, throttle, and brake) and status (speed) in JSON format for training autonomous driving models \cite{bojarski2016end}. 
  \item \textbf{World Status:} Information about the world, such as the weather preset in JSON format for filtering the dataset. 
\end{itemize}

The semantic segmentation GT labels are extracted directly within the Unreal Engine 4 rendering pipeline through the custom stencil buffer after configuring each individual or group of objects in the scene. For generating 2D bounding box annotations for object detection, we heavily depended on the one-hot encoded masks extracted from the semantic segmentation GT label maps for individual or merged classes. The output of the semantic lidar sensor was also used to account for the occlusion of the objects and the drawing of bounding boxes for static meshes that are not directly supported by the CARLA simulator.

\subsubsection{Scenario Simulations}
Inspired by the CARLA simulator scenario runner plug-in\footnote{\url{https://github.com/carla-simulator/scenario_runner}}, we provide a simplified system for spawning and simulating scenarios. This system is parameterizable through YAML configurable files that describe a specific scenario, which can be as simple as spawning the ego vehicle at a random or specific location in the world. More intricate scenarios between the ego vehicle and another actor, such as a pedestrian or a vehicle, can also be simulated through a distance threshold that triggers the controls of the involved actor. Examples of such scenarios are a vehicle that violates a red light or a pedestrian that tries to cross the road, thus becoming a moving obstacle to the ego vehicle.

\subsubsection{Autonomous Driving}
The implementation supports the integration and real-time evaluation of deep-learning methods related to autonomous driving. These include semantic segmentation, object detection, reinforcement learning, and general architectures that predict the controls \cite{bojarski2016end} of an autonomous vehicle. This was achieved through the development of distinct Python algorithms with predefined adaptable methods that are called from the main pipeline (Fig. \ref{fig:flowchart}), which infer with the EPE \cite{Richter_2021} model and control all the functionalities of our tool. Therefore, the approach provides an easy-to-use structure for experimental setups on both the original and the enhanced output of the CARLA rendering pipeline.

\subsubsection{Compilers \& Mixed Precision}

To accelerate the performance of our approach, key tools in this domain, including the SotA TensorRT\footnote{https://developer.nvidia.com/tensorrt/} and ONNX Runtime\footnote{https://onnxruntime.ai/}, are integrated into the implementation. This allowed us to achieve our real-time goal by simultaneously utilizing multiple precision data types (FP32, FP16, TF32, and INT8) to reduce the size of the model and maintain an adequate level of quality. The latter is feasible due to the mixed precision capabilities that are supported by those tools as a result of the latest GPU architectures\footnote{https://developer.nvidia.com/blog/nvidia-ampere-architecture-in-depth/}. Precisely, for the INT8 precision, we included in the approach a calibration procedure that employs a small representative dataset in order to calculate the scale factors for 8-bit inference and avoid the significant accuracy impact due to the low precision. As illustrated in Fig. \ref{fig:fps_comparison}, our implementation can achieve 13 FPS on an RTX 4090 GPU by employing NVIDIA's TensorRT compiler combined with FP16 precision. During experimentation, we observed that applying an autonomous driving task with a low sampling rate did not alter the task performance while significantly improving the FPS of the tool. Therefore, we introduced a mechanism that can skip a number of the simulation steps to enhance the real-time performance.

\begin{figure}
    \centering
    \begin{tikzpicture}[scale=0.9]
        \begin{axis}[
            ybar,
            bar width=0.3cm,
            ylabel={Frames Per Second (FPS)},
            y label style={at={(0.07,0.5)}},
            symbolic x coords={FP32, FP16, FP32-Skip(3), FP16-Skip(3)},
            xtick=data,
            xticklabels={FP32, FP16, FP32-Skip(3), FP16-Skip(3)},
            legend style={at={(0.5,-0.11)}, anchor=north, legend columns=-1},
            legend entries={PyTorch, ONNX Runtime, TensorRT},
            grid=major,
        ]
        
        % Example data points for each method
        \addplot coordinates {(FP32, 2.15) (FP16, 3.13) (FP32-Skip(3), 9.5) (FP16-Skip(3), 10)};
        \addplot coordinates {(FP32, 2.6) (FP16, 3.2) (FP32-Skip(3), 9.15) (FP16-Skip(3), 10.5)};
        \addplot coordinates {(FP32, 3.05) (FP16, 4.6) (FP32-Skip(3), 10.5) (FP16-Skip(3), 13)};
        
        \end{axis}
    \end{tikzpicture}
    \caption{Comparison of the achieved performance of PyTorch \cite{NEURIPS2019_9015}, ONNX Runtime, and TensorRT for FP32 and FP16 precision data types in synchronous mode.}
    \label{fig:fps_comparison}
\end{figure}

\subsection{Limitations} \label{limit_section}

CARLA, being an open-source project, has primarily relied on free or lower-quality assets. However, the introduction of the Town10HD environment signifies a substantial improvement in asset quality. G-Buffers, which provide information about the materials of virtual objects, are particularly informative in this context. Unreal Engine 4 requires only the base color texture of the materials of a virtual object in order to compile and execute the scene. This leads to insufficient information on G-Buffers, where their corresponding textures are not set in the material properties.

This lack of information is highly evident in older CARLA environments, where the employed assets are low-poly, characterized by poor quality, and usually contain little or no information in some specific G-Buffers. For example, the content of the metallic G-Buffer in GTAV \cite{Richter_2021} is sufficiently rich, and its distribution is uniform between the assets, whereas the corresponding one generated from CARLA (Fig. \ref{fig:carla_g_buffers}) provides just different levels of glossiness between each vehicle type (mesh). Along those lines, G-Buffers E (precomputed shadow factors) and F (world-space tangent) were excluded from the synthetic dataset since they were entirely composed of black pixels through all the towns and assets of the CARLA simulator.

In contrast to the GTAV engine, Unreal Engine 4 (UE4) does not provide any dedicated G-Buffers specifically linked to the sky class \cite{Richter_2021}, nor does it offer a buffer related to lighting information, such as emission. The only buffer that provides useful information for the sky class and lighting of the scene is the SceneColor buffer. The lack of a sky-related buffer can be considered a limitation that reduces the effectiveness of that aspect of the environment in comparison with the enhancement results on the Playing for Data (GTAV) dataset \cite{Richter_2021} \cite{Richter_2016_ECCV}.

Moreover, while the approach supports asynchronous mode, as the data synchronization is not guaranteed, there are some specific cases, e.g., when the vehicle steers with high velocity, that lead to ghosting near objects that are drawn having the sky as a background. This is mainly due to the desynchronization of the semantic segmentation label map by a single frame with respect to the other data. This behavior did not alter the performance of the tasks under study, but it is still considered a visual limitation. It was also observed that the asynchronous mode is only reliable for visual evaluation of a task output due to the possibility of desynchronized data.

Eventually, the high hardware requirements of the EPE method were the limiting factor in implementing a multiple-camera configuration for the ego vehicle, as even a single camera poses performance challenges. While compilers such as TensorRT and ONNX Runtime significantly improved the inference, performance limitations persist since the G-Buffer encoder is sensitive to lower-precision data types. The latter led to minimal improvement for INT8 precision and visual artifacts for ONNX Runtime in half-precision.

\begin{figure*}[h]
    \centering
    % First row
    \subfloat[CARLA Town10HD translated towards Cityscapes.]{%
        \label{fig:carla2cs}
        \includegraphics[width=.49\linewidth]{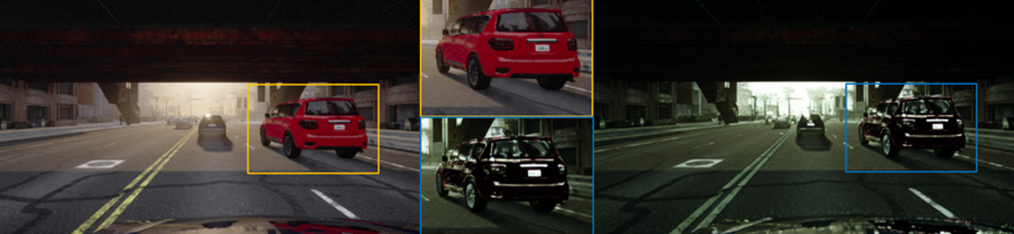}} \hfill
    \subfloat[CARLA Town01 translated towards Cityscapes.]{%
        \label{fig:carla2cs2}
        \includegraphics[width=.475\linewidth]{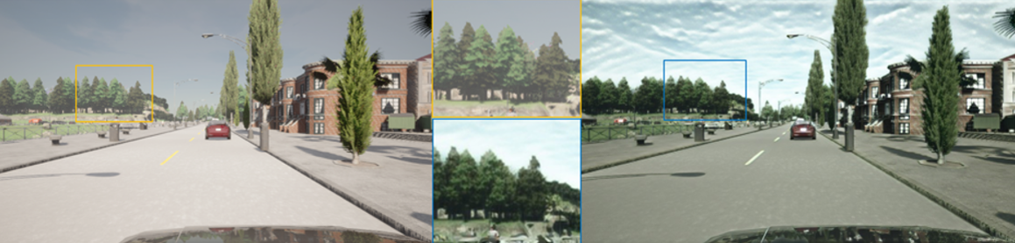}}
    
    \vspace{0.1cm}

    % Second row
    \subfloat[CARLA Town01 translated towards KITTI.]{%
        \label{fig:carla2kitti}
        \includegraphics[width=.49\linewidth]{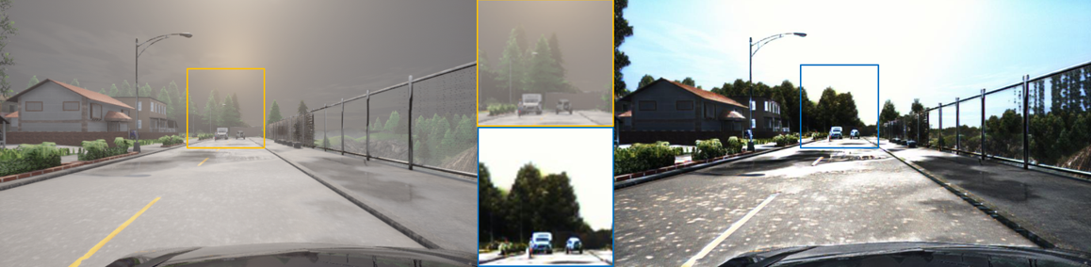}} \hfill
    \subfloat[CARLA Town05 translated towards Mapillary Vistas.]{%
        \label{fig:carla2mv}
        \includegraphics[width=.49\linewidth]{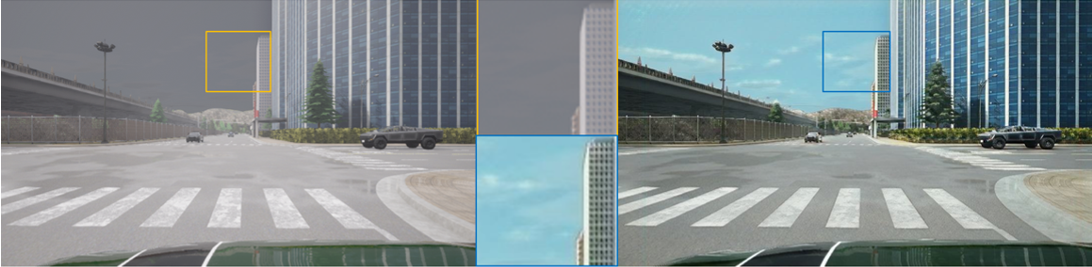}}

    \caption{Translation results of various CARLA environments towards real-world datasets: original CARLA output (left),  the corresponding enhanced/translated output (right).}
%  \label{fig:enhancement_results}
    \label{fig:enhancement_results}
\end{figure*}

\subsection{Translation Results}
\label{trans}
Despite the differences in quality between Town10HD and all older CARLA towns, the results (Fig. \ref{fig:enhancement_results}) seem to provide adequate quality image translations in all cases. As will be illustrated later in the experiments, translating a town that is of low quality and thus has a wider sim2real appearance gap can be even more beneficial compared to the translation of a high-quality environment. Primarily, the model attempts to modify the materials of the objects, making them more realistic, glossy, and closer to the real world. There are also cases in which it tries to alter the geometry of the objects. This is restricted to the vegetation class, where the trees that are of low polygon count and triangular are, after the translation, more densely covered with leaves, appearing a lot more realistic. Furthermore, the grass texture exhibits a more three-dimensional appearance. In the sky class, for most weather presets, the model draws clouds similar to the ones in the real dataset it was trained on, while the color distribution gets closer to the real dataset. Thus, for KITTI, the sky is really bright, while for Cityscapes, the colors are darker with more gray clouds. Additionally, the lighting, which significantly contributes to the realism in a simulation environment and is a part that CARLA lacks, is improved to a high degree, mimicking the lighting of real-world datasets. A significant portion of the translation refers to mimicking the characteristics of the camera that was used to capture the real dataset. The brighter and more vivid colors of Mapillary Vistas are also present in the translated CARLA frames, especially when focusing on the sky. Another limitation of simulations is that the output is often pixelated (object shadows or object edges), which is attributed to the anti-aliasing. This is smoothed by the enhancement model and significantly contributes to realism improvement. However, since the KITTI dataset includes frames that are pixelated due to the camera that was used, in that particular translation result, the output continues to mimic this behavior. The classes that seem to change more (vegetation, vehicle, and road) align with the ones that Richter et al. \cite{Richter_2021} stated when they trained the model with the Playing for Data dataset \cite{Richter_2016_ECCV}. 

\section{Experimental Dataset} \label{Experimental Dataset}

In order to evaluate the impact of the introduced approach on the reduction of the sim2real appearance gap of the CARLA simulator output when deployed in the real world, a number of experiments were conducted. These experiments targeted autonomous driving-related models when trained on the original and the enhanced CARLA data. Eventually, a dataset was exported by employing the proposed tool's dataset generation functionality, targeting a model trained on Cityscapes. The latter was chosen for its superior translation results, as illustrated in Fig. \ref{fig:enhancement_results}, and its close compatibility with CARLA.

The experimental dataset was produced directly through the CARLA simulator in synchronous mode by keeping one in  $20$ frames of the simulation. To avoid any potential bias during the experiments, the original CARLA and the enhanced data shared the same informational content and thus, identical semantic segmentation annotations. Furthermore, considering that the Cityscapes dataset, which consists of $500$ validation frames and GT annotations, was employed as the basis for evaluation during the experiments, a comparable-sized validation set was also extracted for the synthetic data.

Considering that training deep learning models effectively requires a large amount of data to achieve a high degree of robustness, the methodology followed the approach of transfer learning. In more detail,   models based on pre-trained weights of the COCO \cite{cocodataset} (DeepLabV3), or ImageNet \cite{deng2009imagenet} (VGG-16, ResNet-50, and Xception) datasets, depending on the target architecture, were fine-tuned on the experimental dataset. Due to the latter approach, a training set of just 5.000 images was extracted from two distinct CARLA towns, Town10HD and Town01. These differ considerably in realism as a result of the difference in asset quality, which favors the newer, high-definition (Town10HD) town. Thus, the selection of these two towns helped us evaluate the impact of the pre-existing realism gap and its influence on the enhancement effectiveness, thus quantifying the extent to which it alleviates the sim2real appearance gap in the task at hand. To improve the diversity of the dataset, different weather condition presets of the CARLA simulator that align with the daytime scenarios and medium weather conditions of Cityscapes were employed during the generation procedure.

\begin{figure}[htbp]
    \centering
    \begin{overpic}[width=0.5\textwidth,grid=false]{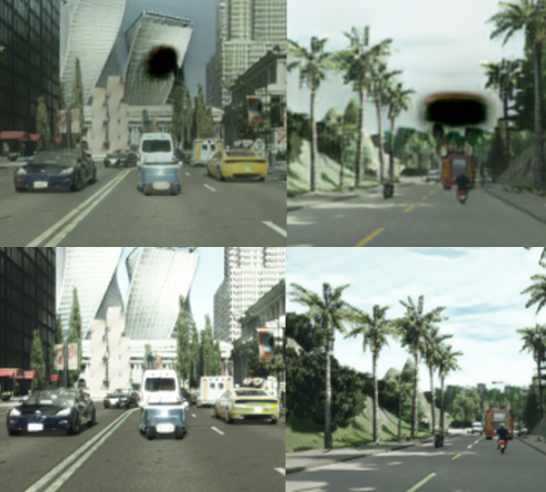}
    \end{overpic}
    
    \caption{Comparison of the translation results of VSAIT (top) compared to the ones produced by CARLA2Real (bottom) in both the Town10HD (left) and Town01 (right) environments.}
    \label{fig:vsait_vs_epe}
\end{figure}

Furthermore, since we are the first to introduce a publicly available dataset that includes additional information generated from the game engine and can be employed for training EPE, the dataset outlined in this section was fed through VSAIT in order to provide extensive quantitative comparisons. Specifically, the official VSAIT code\footnote{https://github.com/facebookresearch/vsait} and parameters were employed along with the frames of the proposed CARLA dataset (Section \ref{synthetic_dataset}) to train a model that can translate CARLA frames towards the characteristics of Cityscapes. Fig. \ref{fig:vsait_vs_epe} illustrates the translation results produced through VSAIT compared with CARLA2Real. As depicted, VSAIT produces few visual artifacts and completely misses the glossiness of the vehicles due to the lack of the related G-Buffers, such as the metallic buffer. Additionally, it is observed that with our proposed dataset and training strategy, the palm trees of Town01 are preserved without any visual artifact, as opposed to the results of Richter et al. \cite{Richter_2021,zheng2022vsait} on the GTA images \cite{Richter_2016_ECCV}.

\section{Feature Analysis}

The proposed tool alters only the output of the camera by adjusting it to render more photorealistic images. On the other hand, a common approach in autonomous driving research, including the SotA algorithms \cite{shao2022interfuser, shao2023reasonnet} featured in the CARLA leaderboard, is to utilize not only the RGB camera but also other sensors, namely LIDAR, RADAR, and IMU. Since the output of these sensors remains unaffected by our enhancement, while playing an important role in the final outcome of the respective algorithm, we opted to perform experiments that are related only to the extracted RGB frames. Specifically, prevalent among these algorithms is the application of a pre-trained feature extraction method \cite{kingma2022autoencoding, simonyan2015deep, he2015deep, xception} to extract features for utilization by a Reinforcement Learning (RL) algorithm.

\subsection{Experimental Setup}
Various approaches and metrics exist for evaluating the similarity between different datasets, including statistical comparisons at the feature level (e.g., cosine similarity \cite{guo-etal-2023-bridging}), distributional comparisons (through e.g., FID and Kernel Inception Distance (KID) \cite{10086509, Richter_2021}), and dimensionality reduction combined with visualization \cite{9915679}. The experimental setup outlined in this section is inspired by \cite{9915679}, where the authors performed a comparison by adopting a dimensionality reduction approach between the embeddings extracted from synthetic datasets (IDDA \cite{Alberti_2020} and Virtual KITTI \cite{gaidon2016virtual}) and the enhanced GTAV images that Richter et al. \cite{Richter_2021} provided. In contrast with \cite{9915679}, our approach is to utilize a metric (cosine similarity) that is applied to the initial feature space (without dimensionality reduction) to support our findings and an explainable AI (XAI) method for results explainability. Additionally, we employ the same datasets before and after applying EPE (CARLA2Real) and VSAIT (Section \ref{Experimental Dataset}). This is because, as it will be demonstrated, a visual gap can also exist between the synthetic datasets, thus resulting in an erroneous evaluation of the actual contribution of the image-to-image translation methods in the reduction of the sim2real appearance gap.

To evaluate the feature similarity of the synthetic datasets compared to the real world, conventional classification architectures, namely VGG-16, ResNet-50, and Xception, integrated into the PyTorch framework, were employed. The selected architectures were fine-tuned using a model pre-trained on the ImageNet \cite{deng2009imagenet} dataset. The classification objective was to predict (classify) the dataset from which a given input frame most likely originated from a total of 7 different datasets (Town01, CARLA2Real Town01, VSAIT Town01, Town10HD, CARLA2Real Town10HD, VSAIT Town10HD, and Cityscapes). This involved the Adam optimizer with a learning rate of 0.0001, cross-entropy loss, and a selection process to retain the best-performing model based on the validation accuracy. The preprocessing steps were split into two phases. First, the frames of all datasets were resized to 512x256, and the hood of the ego vehicle was removed to eliminate any potential bias. Second, standard preprocessing steps were applied. These included center cropping of the images to match the required input resolution of the models and application of mean and standard deviation normalization required for each architecture. Additionally, we employed the GRAD-CAM XAI \cite{gradcam} method to investigate how the photorealism enhancement method utilized by CARLA2Real influences the decision-making process.

During the evaluation phase, we utilized for each of the synthetic datasets (classes) before and after applying the image-to-image translation methods, namely Town01, CARLA2Real Town01, VSAIT Town01, Town10HD, CARLA2Real Town10HD, and VSAIT Town10HD, a validation set that contained 500 frames, matching the size of the Cityscapes validation set. This was done in order to extract features from the last convolutional layer of each architecture (VGG-16, Resnet-50, and Xception) and compare their pairwise similarity with the ones extracted from the real-world dataset (Cityscapes). In more detail, these features were extracted from both the pretrained (on ImageNet) weights and after fine-tuning each of the architectures for each dataset type, resulting in six feature comparisons, namely, CARLA Town01-Cityscapes, CARLA Town10HD-Cityscapes, CARLA2Real Town01-Cityscapes, CARLA2Real Town10HD-Cityscapes, VSAIT Town01-Cityscapes, and VSAIT Town10HD-Cityscapes. To evaluate similarity without dimensionality reduction and potential information loss, pairwise cosine similarity was employed in the original high-dimensional space. Additionally, to avoid bias in the comparisons, the synthetic and enhanced datasets contained identical images with only the applied modifications of the CARLA2Real and VSAIT approaches, which prevent semantic flipping.

\begin{table*}[htbp]
\footnotesize
\caption{\label{fs2}Feature similarity analysis of the features extracted from various CARLA dataset variations compared to those extracted from Cityscapes. The pairwise cosine similarity metric is used. Bold values highlight the highest similarity for each CARLA town. 'PT' refers to the pretrained models, while 'FT' indicates the same models after fine-tuning.}
\begin{center}
\begin{tabular}{|c|c|c|c|c|c|c|}
\hline
\textbf{Dataset} & \textbf{ResNet50 (FT)} & \textbf{ResNet50 (PT)} & \textbf{VGG-16 (FT)} & \textbf{VGG-16 (PT)} & \textbf{Xception (FT)} & \textbf{Xception (PT)} \\
\hline
\textit{Town10HD} & 0.5134 & 0.7822 & 0.1716 & 0.2512 & 0.3514 & 0.5692 \\
\textit{CARLA2Real Town10HD} & \textbf{0.6128} & \textbf{0.7961} &  \textbf{0.2531} &  \textbf{0.2636} &  \textbf{0.4570} &  \textbf{0.5876} \\
\textit{VSAIT Town10HD} & 0.5213 & 0.7788 & 0.2052 & 0.2165 & 0.3982 & 0.5536 \\
\hdashline
\textit{Town01} &  0.4838  & 0.7571 & 0.1734 & 0.2027 & 0.2506 & 0.5333 \\
\textit{CARLA2Real Town01} & \textbf{0.5637} & \textbf{0.7780} & 0.2284 & 0.2198 & \textbf{0.3910} & 0.5474 \\
\textit{VSAIT Town01} & 0.4930 & 0.7700 &  \textbf{0.2721} &  \textbf{0.2308} & 0.3563 &  \textbf{0.5673} \\
\hline
\end{tabular}
\end{center}
\end{table*}

\subsection{Results and Discussion}

Table \ref{fs2} presents the pairwise cosine similarity values obtained from comparing a.1) features extracted from simulator-generated images and a.2) simulator images upon which EPE (CARLA2Real) and VSAIT approaches (targeting Cityscapes) were applied with b) features from the Cityscapes dataset. It is observed that similarity with Cityscapes features increases when utilizing features extracted from the translated data. This is also illustrated in the case of utilizing the pre-trained (PT columns in Table \ref{fs2}) models on ImageNet \cite{deng2009imagenet}, yet the increase is less significant, since the models are not specifically fine-tuned for that task and tend to treat the features similarly.

Furthermore, as observed in Table \ref{fs2}, the features derived from the translated data produced by CARLA2Real achieve a higher cosine similarity with Cityscapes compared to the ones generated by VSAIT across the majority of models (pre-trained and fine-tuned) and CARLA Towns. This can be attributed to the small visual artifacts that were observed on the VSAIT results (Fig. \ref{fig:vsait_vs_epe}) and the fact that these results are not as good, in terms of, e.g., enhanced glossiness, as those achieved by EPE (CARLA2Real) since the latter utilizes the G-Buffers. In detail, VSAIT outperforms CARLA2Real in just a few instances in the Town01 environment, possibly because it generates fewer artifacts in this town.

Additionally, when looking at the Town10HD and Town01 rows in Table \ref{fs2}, it is apparent that a sim2real appearance gap can exist in the simulator itself. Indeed, the features extracted from the Town10HD environment achieve a higher similarity with the ones derived from the real-world frames of Cityscapes compared to Town01. This is due to the significantly improved quality of the assets that were employed in Town10HD. However, results show that both methods (CARLA2Real and VSAIT) can significantly improve the visual realism of the lower- and higher-quality CARLA towns. In detail, the translated data of Town01 (CARLA2Real Town01) exhibit a lower feature similarity with Cityscapes compared to the ones extracted from the enhanced images of Town10HD (CARLA2Real Town10HD) due to the initial baseline difference. The official release of the Unreal Engine 5 version of the CARLA simulator will mitigate this since all the towns will employ visually upgraded assets.

\begin{figure}[htbp]
    \centering
    \begin{overpic}[width=0.5\textwidth,grid=false]{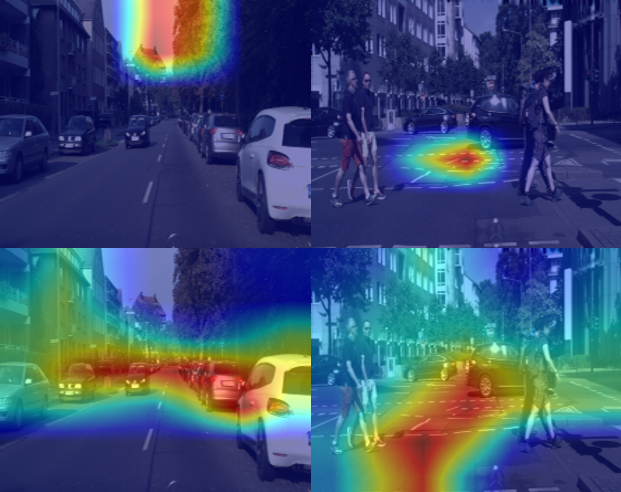}
    \end{overpic}
    
    \caption{Grad-CAM heatmaps visualizations demonstrating the Resnet-50 model's focus on two Cityscapes validation frames when predicting the probability of the CARLA (top) and CARLA2Real (bottom) classes.}
    \label{fig:gradcam_visualizations}
\end{figure}

Finally, the GRAD-CAM visualization applied on the ResNet-50 architecture (Fig. \ref{fig:gradcam_visualizations}) illustrates that the model focuses mainly on certain, limited parts of the image when predicting the simulator-related classes (Town10HD and Town01). In contrast, the CARLA2Real classes (CARLA2Real Town10HD and CARLA2Real Town01) indicate a more extensive examination of a wider variety of areas in the input image during decision-making. These include people, vegetation, roads, and cars. This observation suggests that in a situation where the sim2real appearance gap is substantial, the model confidently makes decisions by focusing on a limited image region. However, as the sim2real appearance gap shrinks due to the enhancement, the model encounters increased difficulty in distinguishing between CARLA2Real and Cityscapes images.

\section{Semantic Segmentation}

Semantic segmentation is a computer vision task that is closely related to autonomous driving, as it provides a pixel-wise understanding of the vehicle's environment. This level of detailed scene understanding enables the vehicle to avoid collisions and react to objects in the environment, such as traffic lights, signs, and pedestrians. The majority of the methods currently utilized for semantic segmentation \cite{chen2017rethinking, chen2018encoderdecoder, he2018mask, ronneberger2015unet} leverage a pre-trained convolutional neural network (CNN), such as ResNet, at the initial stages of the network architecture to extract features from the input frame. These backbones \cite{simonyan2015deep, xception, he2015deep} have been analyzed in the previous section with respect to the feature similarity between the original and the enhanced data compared to the features extracted from real-world images. In this section, we investigate whether the increase in the similarity of features evaluated on enhanced data with those derived from real data can influence semantic segmentation accuracy.

\subsection{Experimental Setup}

Google’s DeeplabV3 \cite{chen2017rethinking} using a Resnet-50 \cite{he2015deep} backbone was employed in the semantic segmentation experimentation, since it is integrated into the PyTorch framework \cite{NEURIPS2019_9015}. The pre-trained weights originate from a model that was trained on a subset of COCO \cite{cocodataset} train 2017, specifically on the 20 categorical classes of the PASCAL VOC \cite{pascal-voc-2012} dataset. Standard DeeplabV3 preprocessing procedures were followed, namely image loading within the [0,1] range, followed by normalization using mean values of [0.485, 0.456, 0.406] and standard deviations of [0.229, 0.224, 0.225]. To augment the data and improve the robustness of the model, random horizontal and vertical flips were applied before selecting a random 224x224 crop of the full-resolution input image. The training lasted 20 epochs, utilizing the Adam optimizer with a learning rate of 1e-4, a cross-entropy loss function, and a batch size of 8.

Once training was over, the best-performing model was selected based on the average Intersection over Union (IoU) of the validation set, which consists of unseen data. The training was carried out using the training set from the Cityscapes dataset and synthetic datasets obtained from Town10HD and Town01. This allowed the assessment of the variations in quality and translation effectiveness between the older Town01 town and the recently developed high-definition Town10HD.

The evaluation phase included the assessment of four distinct types of models trained on the original CARLA, CARLA2Real, VSAIT, and the real-world frames of Cityscapes. All trained models were evaluated on the Cityscapes validation set, as the test set does not contain GT annotations for calculating the IoU. In addition, the CARLA-trained models were also evaluated on their respective validation sets. This was performed in order to estimate whether the enhanced data can further improve the accuracy of Deeplabv3 when deployed strictly inside the simulation environment. In addition, all types of trained models were cross-evaluated with the validation sets of all the dataset types. In this case study, a significant drop between the models and data would indicate that the data has changed to the degree that the model no longer considers them to be the same. The same preprocessing procedure with the training phase was applied, except for the augmentation steps that were discarded during model evaluation. In this experimental phase, the data were passed through the model with a batch size of one, as in CARLA, which is applied to each frame of the simulator. The average IoU for each frame, considering all classes and per-class accuracy, is computed after iterating through all the frames of the validation set.

\begin{table}[ht]
\footnotesize
\centering
\caption{\label{sem_seg_table_all_1}Evaluation of Town10HD DeepLabV3-trained models, comparing performance across different dataset types using Intersection over Union (IoU) as the evaluation metric.}
\begin{tabular}{|c|c|c|c|c|}
\hline
\textbf{Trained Model} & \multicolumn{4}{|c|}{\textbf{Test Set}} \\ 
\hline
 & \textbf{\textit{CARLA}} & \textbf{\textit{CARLA2Real}} & \textbf{\textit{VSAIT}} & \textbf{\textit{Cityscapes}} \\
\hline
CARLA & 56.23 & 32.78 & 13.58 & 16.48 \\
\hline
CARLA2Real & 31.90 & 55.42 & 44.77 & 30.21 \\
\hline
VSAIT & 38.12 & 49.25 & 54.31 & 26.91 \\
\hline
Cityscapes & 28.22 & 34.55 & 31.18 & 51.50 \\
\hline
\end{tabular}
\end{table}

\begin{table}[ht]
\footnotesize
\centering
\caption{\label{sem_seg_table_all_2}Evaluation of Town01 DeepLabV3-trained models, comparing performance across different dataset types using IoU as the evaluation metric.}
\begin{tabular}{|c|c|c|c|c|}
\hline
\textbf{Trained Model} & \multicolumn{4}{|c|}{\textbf{Test Set}} \\
\hline
 & \textbf{\textit{CARLA}} & \textbf{\textit{CARLA2Real}} & \textbf{\textit{VSAIT}} & \textbf{\textit{Cityscapes}} \\
\hline
CARLA & 62.32 & 22.45 & 20.23 & 14.71 \\
\hline
CARLA2Real & 43.70 & 62.71 & 52.10 & 31.92 \\
\hline
VSAIT & 52.10 & 58.60 & 61.62 & 27.47 \\
\hline
Cityscapes & 34.69 & 43.99 & 37.89 & 51.50 \\
\hline
\end{tabular}
\end{table}

\begin{table}[ht]
\footnotesize
\centering
\caption{\label{sem_seg_table_per_class}Per-class evaluation of DeepLabV3 models trained on (a) Town10HD (original and enhanced), (b) VSAIT, and (c) Cityscapes on the Cityscapes validation set. Higher is better. Bold values indicate the most significant increase in accuracy after applying the photorealism enhancement approach. IoU is used as the evaluation metric.}
\begin{tabular}{|c|c|c|c|c|}
\hline
\textbf{Class} & \multicolumn{4}{|c|}{\textbf{Trained Model}} \\
\hline
& \textbf{\textit{CARLA}} & \textbf{\textit{CARLA2Real}} & \textbf{\textit{VSAIT}} & \textbf{\textit{Cityscapes}} \\
\hline
Background & 1.89 & 8.92 & 11.21 & 23.71 \\
\hline
Sidewalk & 0.65 & \textbf{21.80} & 15.09 & 43.42 \\
\hline
Road & 27.77 & \textbf{84.94} & 76.69 & 91.60 \\
\hline
Person & 0.61 & \textbf{6.38} & 3.13 & 28.23 \\
\hline
Vegetation & 48.51 & \textbf{67.93} & 55.55 & 80.09 \\
\hline
Building & 25.14 & \textbf{56.24} & 50.57 & 70.85 \\
\hline
Cars & 33.30 & 38.10 & 44.28 & 84.30 \\
\hline
Sky & 47.75 & \textbf{60.11} & 58.80 & 75.86 \\
\hline
Bus & 1.24 & 3.69 & 3.10 & 33.80 \\
\hline
Traffic Sign & 1.99 & 2.38 & 3.39 & 25.02 \\
\hline
Traffic Light & 1.55 & 1.49 & 2.12 & 15.84 \\
\hline
\end{tabular}
\end{table}

\begin{table}[ht]
\footnotesize
\centering
\caption{\label{sem_seg_table_per_class_2}Per-class evaluation of DeepLabV3 models trained on (a) Town01 (original and enhanced), (b) VSAIT, and (c) Cityscapes on the Cityscapes validation set. Higher is better. Bold values indicate the most significant increase in accuracy after applying the photorealism enhancement approach. IoU is used as the evaluation metric.}
\begin{tabular}{|c|c|c|c|c|}
\hline
\textbf{Class} & \multicolumn{4}{|c|}{\textbf{Trained Model}} \\
\hline
& \textbf{\textit{CARLA}} & \textbf{\textit{CARLA2Real}} & \textbf{\textit{VSAIT}} & \textbf{\textit{Cityscapes}} \\
\hline
Background & 5.34 & 13.33 & 9.92 & 23.71 \\
\hline
Sidewalk & 1.11 & \textbf{31.66} & 25.53 & 43.42 \\
\hline
Road & 6.41 & \textbf{76.15} & 69.65 & 91.60 \\
\hline
Person & 4.22 & \textbf{8.27} & 5.24 & 28.23 \\
\hline
Vegetation & 42.87 & \textbf{68.46} & 66.01 & 80.09 \\
\hline
Building & 27.93 & \textbf{53.40} & 46.33 & 70.85 \\
\hline
Cars & 26.61 & \textbf{43.06} & 34.97 & 84.03 \\
\hline
Sky & 43.62 & \textbf{56.28} & 50.80 & 75.86 \\
\hline
Bus & 0.8 & 0.62 & 1.95 & 33.80 \\
\hline
Traffic Sign & 0.63 & 1.60 & 0.96 & 25.02 \\
\hline
Traffic Light & 0.3 & 1.58 & 1.2 & 15.84 \\
\hline
\end{tabular}
\end{table}

\subsection{Results and Discussion}

The results presented in Tables \ref{sem_seg_table_all_1} and \ref{sem_seg_table_all_2} of this section demonstrate a significant decline in segmentation accuracy when evaluating a model trained on the photorealism-enhanced output (CARLA2Real and VSAIT) on the original CARLA validation set (column CARLA in Tables \ref{sem_seg_table_all_1} and \ref{sem_seg_table_all_2}). This decline indicates that the photorealism enhancement method introduced a sufficient level of dissimilarity (domain shift) to the degree that the model fails to generalize to the original rendered images. Particularly, the domain shift introduced by CARLA2Real is more significant compared to VSAIT, which does not rely on G-Buffers and fails to enhance various properties of the scene, such as the glossiness of the vehicles, while being more prone to visual artifacts. In addition, the results show that training a model in order to be deployed in the enhanced simulator (training on the CARLA2Real or VSAIT dataset and testing on the validation set of the same dataset) did not yield any additional gains in comparison with the original CARLA-trained model tested on its corresponding validation set. This leads to the conclusion that training a model exclusively for deployment inside the enhanced simulator lacks any significant advantages compared to the conventional models trained on the original simulator output. Moreover, it is seen that the models trained on the data produced by CARLA2Real  demonstrated a significant increase in performance compared to the models trained on the original CARLA data when both are evaluated on the real-world Cityscapes data. This increase is more significant than the one introduced by VSAIT, as can be seen when inspecting the Cityscapes column in Tables \ref{sem_seg_table_all_1} and \ref{sem_seg_table_all_2}, and aligns with the weaker domain shift introduced by VSAIT. The same applies in the case of evaluating the Cityscapes-trained model on the enhanced data (CARLA2Real and VSAIT) compared to the original CARLA frames (Cityscapes row in Tables \ref{sem_seg_table_all_1} and \ref{sem_seg_table_all_2}).

\begin{figure}[htbp]
    \centering
    \begin{overpic}[width=\textwidth,grid=false]{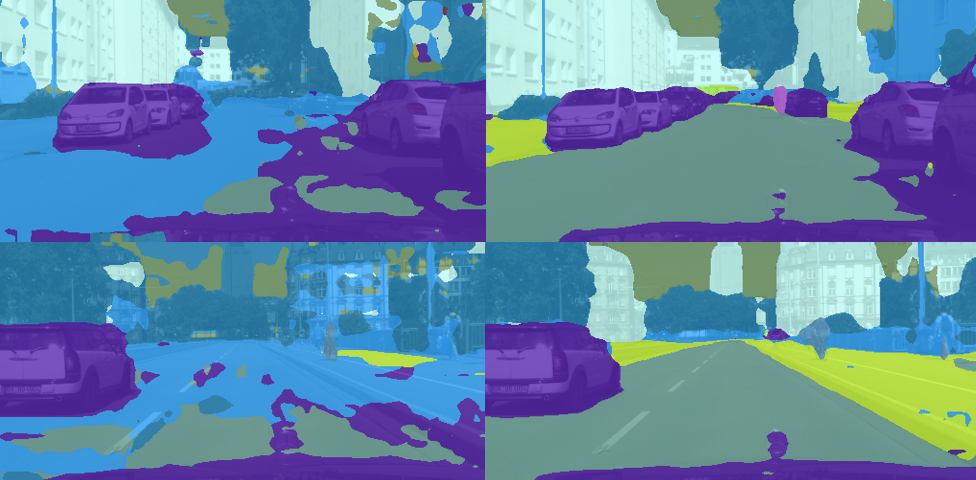}
    \end{overpic}
    
    \caption{Semantic segmentation results obtained from a CARLA-trained model (left) and a CARLA2Real-trained model (right) on Cityscapes validation frames.}
    \label{fig:sem_seg_cs_predictions}
\end{figure}

To further investigate these observations, we refer to Tables \ref{sem_seg_table_per_class} and \ref{sem_seg_table_per_class_2}, which provide the per-class accuracy of the experiments and reveal similar findings. It is apparent that, when training with the CARLA2Real data and testing on Cityscapes, accuracy more than doubles for some specific classes compared to when performing the same experiment on the original data (columns CARLA and CARLA2Real in Tables \ref{sem_seg_table_per_class} and \ref{sem_seg_table_per_class_2}). This shows that the enhanced data produced by CARLA2Real can indeed reduce the appearance gap of the Unreal Engine rendering pipeline and achieve higher performance when utilized by semantic segmentation models that are to be deployed in the real world. Indeed, a model trained on CARLA2Real enhanced data produces superior semantic segmentation accuracy in comparison to the original simulator output, while outperforming VSAIT on most autonomous driving-related classes, such as road, sidewalk, and person. Additionally, the proposed tool improves the semantic segmentation accuracy despite the significant visual gap between the CARLA Town10HD and Town01 environments. Both towns benefit from nearly a twofold increase in IoU when the trained DeepLabV3 models are evaluated on the Cityscapes validation set (column Cityscapes in Tables \ref{sem_seg_table_all_1} and \ref{sem_seg_table_all_2}). These observations are also exemplified in Fig. \ref{fig:sem_seg_cs_predictions}, which provides a visual representation of the semantic segmentation predictions made on two sample frames of the Cityscapes validation set by a CARLA-trained DeepLabV3 model in comparison with the counterparts trained on the enhanced data produced by CARLA2Real. It is evident that training with the original CARLA simulator data yields results that are inconsistent. On the other hand, the model trained on the data of the introduced approach demonstrates significantly higher accuracy in its prediction. This is particularly true for the road, car, vegetation, sky, and building classes that are simultaneously consistent.

Results in Tables \ref{sem_seg_table_per_class} and \ref{sem_seg_table_per_class_2} (specifically columns CARLA, CARLA2Real, and VSAIT) also show that, when training with the enhanced data, despite the significant percentage improvement in accuracy for some specific classes, the IoU metric remains low compared to the results achieved by the model that was trained on Cityscapes. Since the goal of these image-to-image translation methods (EPE and VSAIT) is to preserve the overall environment structure so as to prevent visual artifacts, this is an anticipated behavior that can be attributed to the content gap that exists between the CARLA and the Cityscapes object distribution.

In more detail, this disparity between CARLA and Cityscapes exists due to the fact that they depict different geographic regions (the USA and Europe, respectively). This difference is particularly prominent in classes such as traffic signs, which differ considerably between the two regions and thus illustrate a really low performance in terms of IoU. In \cite{gta_deeplabv3}, which employed similar image-to-image translation methods that altered the geometry to a higher degree,  significantly low accuracy on region-specific classes such as traffic signs, compared to the Cityscapes-trained models, was also observed.

Additionally, the content gap is not restricted to regional differences but is also significant due to the limited variety and distribution of objects inside the simulator. For example, the bus class in the CARLA simulator contains only a single vehicle\footnote{\url{https://carla.readthedocs.io/en/latest/catalogue_vehicles/}} with completely different characteristics from the corresponding class in Cityscapes, which also exhibits significant diversity. In contrast, for more universal, not region-specific classes, such as vegetation, for which CARLA includes a wide variety of instances, there is a significant improvement in performance. This performance, in some cases, approaches the performance of the models trained on real-world (Cityscapes) data. The classes that proved to significantly benefit from the proposed tool include the vegetation, road, and vehicle classes. This also aligns with the observations of Richter et al. \cite{Richter_2021} that these classes yielded the best translation results during their qualitative experimentation on GTAV \cite{Richter_2016_ECCV}.

\section{Conclusion}
In this paper, a robust approach was employed to enhance the photorealism of the CARLA simulator, towards closing the sim2real appearance gap, to the benefit of autonomous driving-related research. Based on this, the  CARLA2real tool was developed. This publicly available tool optimizes the implementation of such an image-to-image translation model in order to enable its real-time application along with fundamental autonomous driving functionalities. The latter includes synthetic dataset extraction, autonomous driving-related task evaluation, and scenario simulations parametrized through a configuration file. To evaluate the effectiveness of CARLA2real, we performed a number of experiments on feature extraction and semantic segmentation methods that are essential in the domain of autonomous driving. We also compared the effect of the proposed approach with that of a SotA method that does not utilize additional information generated by the game engine. 

The results illustrated that the approach can improve the similarity of the features extracted from the simulation data with features extracted from real-world data and increase the difficulty of the models to distinguish between the enhanced and the real-world images. This improvement in feature similarity also led to an increase in the performance of a semantic segmentation model, namely DeepLabV3. This model reached up to double the accuracy when evaluated on the real-world Cityscapes validation set while outperforming the results achieved by the VSAIT method, which did not employ additional G-Buffers. Yet, the models trained on the enhanced data failed to achieve high accuracy in objects that are characterized by a content gap. Such a behavior was expected since these photorealism enhancement methods do not alter the geometry and content of the objects in their effort to retain the overall structure of the scene and prevent visual artifacts.

As future work, we plan to combine the proposed approach with other methods that focus on closing the sim2real content gap. By doing so, we would like to investigate if their combination with the proposed tool can further shrink the overall sim2real gap in CARLA.

	\paragraph{Acknowledgements} The work presented here has been partially supported by the RoboSAPIENS project funded by the European Union
Horizon Europe programme under grant agreement
no 101133807. 
	
%	\newpage
	\bibliography{refs}

\begin{thebibliography}{63}
\providecommand{\natexlab}[1]{#1}
\providecommand{\url}[1]{\texttt{#1}}
\expandafter\ifx\csname urlstyle\endcsname\relax
  \providecommand{\doi}[1]{doi: #1}\else
  \providecommand{\doi}{doi: \begingroup \urlstyle{rm}\Url}\fi

\bibitem[Alberti et~al.(2020)Alberti, Tavera, Masone, and Caputo]{Alberti_2020}
Emanuele Alberti, Antonio Tavera, Carlo Masone, and Barbara Caputo.
\newblock Idda: A large-scale multi-domain dataset for autonomous driving.
\newblock \emph{\textit{IEEE Robotics and Automation Letters}}, 5\penalty0 (4):\penalty0 5526--5533, 2020.
\newblock \href{https://doi.org/10.1109/LRA.2020.3009075}{\ttfamily\path{ doi:10.1109/LRA.2020.3009075}}.

\bibitem[Bansal et~al.(2016)Bansal, Chen, Russell, Gupta, and Ramanan]{bansal2016pixelnet}
Aayush Bansal, Xinlei Chen, Bryan Russell, Abhinav Gupta, and Deva Ramanan.
\newblock Pixelnet: Towards a general pixel-level architecture, 2016.

\bibitem[Bińkowski et~al.(2018)Bińkowski, Sutherland, Arbel, and Gretton]{bińkowski2018demystifying}
Mikołaj Bińkowski, Dougal~J. Sutherland, Michael Arbel, and Arthur Gretton.
\newblock Demystifying {MMD} {GAN}s.
\newblock In \emph{International Conference on Learning Representations}, 2018.
\newblock URL \url{https://openreview.net/forum?id=r1lUOzWCW}.

\bibitem[Bojarski et~al.(2016)Bojarski, Testa, Dworakowski, Firner, Flepp, Goyal, Jackel, Monfort, Muller, Zhang, Zhang, Zhao, and Zieba]{bojarski2016end}
Mariusz Bojarski, Davide~Del Testa, Daniel Dworakowski, Bernhard Firner, Beat Flepp, Prasoon Goyal, Lawrence~D. Jackel, Mathew Monfort, Urs Muller, Jiakai Zhang, Xin Zhang, Jake Zhao, and Karol Zieba.
\newblock End to end learning for self-driving cars.
\newblock \href{https://arxiv.org/abs/1604.07316}{arXiv:1604.07316}, 2016.

\bibitem[Bujwid et~al.(2018)Bujwid, Martí, Azizpour, and Pieropan]{bujwid2018gantruthunpairedimagetoimage}
Sebastian Bujwid, Miquel Martí, Hossein Azizpour, and Alessandro Pieropan.
\newblock Gantruth - an unpaired image-to-image translation method for driving scenarios, 2018.

\bibitem[Chen et~al.(2017)Chen, Papandreou, Schroff, and Adam]{chen2017rethinking}
Liang-Chieh Chen, George Papandreou, Florian Schroff, and Hartwig Adam.
\newblock Rethinking atrous convolution for semantic image segmentation, 2017.

\bibitem[Chen et~al.(2018)Chen, Zhu, Papandreou, Schroff, and Adam]{chen2018encoderdecoder}
Liang-Chieh Chen, Yukun Zhu, George Papandreou, Florian Schroff, and Hartwig Adam.
\newblock Encoder-decoder with atrous separable convolution for semantic image segmentation.
\newblock In Vittorio Ferrari, Martial Hebert, Cristian Sminchisescu, and Yair Weiss, editors, \emph{\textit{Computer Vision -- ECCV 2018}}, pages 833--851, Cham, 2018. Springer International Publishing.

\bibitem[Chollet(2017)]{xception}
F.~Chollet.
\newblock Xception: Deep learning with depthwise separable convolutions.
\newblock In \emph{\textit{2017 IEEE Conference on Computer Vision and Pattern Recognition (CVPR)}}, pages 1800--1807, Los Alamitos, CA, USA, jul 2017. IEEE Computer Society.
\newblock \href{https://doi.org/10.1109/CVPR.2017.195}{\ttfamily\path{ doi:10.1109/CVPR.2017.195}}.

\bibitem[Cordts et~al.(2016)Cordts, Omran, Ramos, Rehfeld, Enzweiler, Benenson, Franke, Roth, and Schiele]{cordts2016cityscapes}
Marius Cordts, Mohamed Omran, Sebastian Ramos, Timo Rehfeld, Markus Enzweiler, Rodrigo Benenson, Uwe Franke, Stefan Roth, and Bernt Schiele.
\newblock The cityscapes dataset for semantic urban scene understanding.
\newblock In \emph{\textit{Proc. of the IEEE Conference on Computer Vision and Pattern Recognition (CVPR)}}, 2016.

\bibitem[Deng et~al.(2009)Deng, Dong, Socher, Li, Li, and Fei-Fei]{deng2009imagenet}
Jia Deng, Wei Dong, Richard Socher, Li-Jia Li, Kai Li, and Li~Fei-Fei.
\newblock Imagenet: A large-scale hierarchical image database.
\newblock In \emph{\textit{2009 IEEE Conference on Computer Vision and Pattern Recognition}}, pages 248--255, 2009.
\newblock \href{https://doi.org/10.1109/CVPR.2009.5206848}{\ttfamily\path{ doi:10.1109/CVPR.2009.5206848}}.

\bibitem[Dosovitskiy et~al.(2017)Dosovitskiy, Ros, Codevilla, Lopez, and Koltun]{dosovitskiy2017carla}
Alexey Dosovitskiy, German Ros, Felipe Codevilla, Antonio Lopez, and Vladlen Koltun.
\newblock {CARLA}: {An} open urban driving simulator.
\newblock In \emph{\textit{Proceedings of the 1st Annual Conference on Robot Learning}}, pages 1--16, 2017.

\bibitem[Efangelos and Stafylopatis(2022)]{gta_deeplabv3}
T.~Efangelos and A.~Stafylopatis.
\newblock Unsupervised translation of grand theft auto v images to real urban scenes.
\newblock \url{https://dspace.lib.ntua.gr/xmlui/handle/123456789/54709}, 2022.

\bibitem[Elmquist et~al.(2025)Elmquist, Serban, and Negrut]{10819363}
Asher Elmquist, Radu Serban, and Dan Negrut.
\newblock A methodology to quantify simulation- versus-reality differences in images for autonomous robots.
\newblock \emph{IEEE Sensors Journal}, 25\penalty0 (4):\penalty0 6522--6533, 2025.
\newblock \href{https://doi.org/10.1109/JSEN.2024.3522050}{\ttfamily\path{ doi:10.1109/JSEN.2024.3522050}}.

\bibitem[Everingham et~al.(2012)Everingham, Van~Gool, Williams, Winn, and Zisserman]{pascal-voc-2012}
M.~Everingham, L.~Van~Gool, C.~K.~I. Williams, J.~Winn, and A.~Zisserman.
\newblock The {PASCAL} {V}isual {O}bject {C}lasses {C}hallenge 2012 {(VOC2012)} {R}esults, 2012.

\bibitem[Gadipudi et~al.(2022)Gadipudi, Elamvazuthi, Sanmugam, Izhar, Prasetyo, Jegadeeshwaran, and Ali]{9915679}
Nivesh Gadipudi, Irraivan Elamvazuthi, Mahindra Sanmugam, Lila~Iznita Izhar, Tindyo Prasetyo, R~Jegadeeshwaran, and Syed Saad~Azhar Ali.
\newblock Synthetic to real gap estimation of autonomous driving datasets using feature embedding.
\newblock In \emph{\textit{2022 IEEE 5th International Symposium in Robotics and Manufacturing Automation (ROMA)}}, pages 1--5, 2022.
\newblock \href{https://doi.org/10.1109/ROMA55875.2022.9915679}{\ttfamily\path{ doi:10.1109/ROMA55875.2022.9915679}}.

\bibitem[Gaidon et~al.(2016)Gaidon, Wang, Cabon, and Vig]{gaidon2016virtual}
Adrien Gaidon, Qiao Wang, Yohann Cabon, and Eleonora Vig.
\newblock Virtualworlds as proxy for multi-object tracking analysis.
\newblock In \emph{\textit{2016 IEEE Conference on Computer Vision and Pattern Recognition (CVPR)}}, pages 4340--4349, 2016.
\newblock \href{https://doi.org/10.1109/CVPR.2016.470}{\ttfamily\path{ doi:10.1109/CVPR.2016.470}}.

\bibitem[Geiger et~al.(2013)Geiger, Lenz, Stiller, and Urtasun]{liao2022kitti360}
Andreas Geiger, Philip Lenz, Christoph Stiller, and Raquel Urtasun.
\newblock Vision meets robotics: The kitti dataset.
\newblock \emph{\textit{International Journal of Robotics Research (IJRR)}}, 2013.

\bibitem[Goodfellow et~al.(2014{\natexlab{a}})Goodfellow, Pouget-Abadie, Mirza, Xu, Warde-Farley, Ozair, Courville, and Bengio]{goodfellow2014generative}
Ian Goodfellow, Jean Pouget-Abadie, Mehdi Mirza, Bing Xu, David Warde-Farley, Sherjil Ozair, Aaron Courville, and Yoshua Bengio.
\newblock Generative adversarial nets.
\newblock In \emph{\textit{Advances in neural information processing systems}}, pages 2672--2680, 2014{\natexlab{a}}.

\bibitem[Goodfellow et~al.(2014{\natexlab{b}})Goodfellow, Pouget-Abadie, Mirza, Xu, Warde-Farley, Ozair, Courville, and Bengio]{NIPS2014_f033ed80}
Ian~J. Goodfellow, Jean Pouget-Abadie, Mehdi Mirza, Bing Xu, David Warde-Farley, Sherjil Ozair, Aaron Courville, and Yoshua Bengio.
\newblock Generative adversarial nets.
\newblock In Z.~Ghahramani, M.~Welling, C.~Cortes, N.~Lawrence, and K.Q. Weinberger, editors, \emph{Advances in Neural Information Processing Systems}, volume~27. Curran Associates, Inc., 2014{\natexlab{b}}.
\newblock URL \url{https://proceedings.neurips.cc/paper_files/paper/2014/file/f033ed80deb0234979a61f95710dbe25-Paper.pdf}.

\bibitem[Guo et~al.(2023)Guo, Fang, Yu, and Feng]{guo-etal-2023-bridging}
Wenyu Guo, Qingkai Fang, Dong Yu, and Yang Feng.
\newblock Bridging the gap between synthetic and authentic images for multimodal machine translation.
\newblock In \emph{Proceedings of the 2023 Conference on Empirical Methods in Natural Language Processing}, pages 2863--2874, Singapore, 2023. Association for Computational Linguistics.
\newblock \href{https://doi.org/10.18653/v1/2023.emnlp-main.173}{\ttfamily\path{ doi:10.18653/v1/2023.emnlp-main.173}}.

\bibitem[Han et~al.(2021)Han, Shoeiby, Petersson, and Armin]{han2021dual}
Junlin Han, Mehrdad Shoeiby, Lars Petersson, and Mohammad~Ali Armin.
\newblock Dual contrastive learning for unsupervised image-to-image translation.
\newblock In \emph{\textit{2021 IEEE/CVF Conference on Computer Vision and Pattern Recognition Workshops (CVPRW)}}, pages 746--755, 2021.
\newblock \href{https://doi.org/10.1109/CVPRW53098.2021.00084}{\ttfamily\path{ doi:10.1109/CVPRW53098.2021.00084}}.

\bibitem[He et~al.(2016)He, Zhang, Ren, and Sun]{he2015deep}
Kaiming He, Xiangyu Zhang, Shaoqing Ren, and Jian Sun.
\newblock Deep residual learning for image recognition.
\newblock In \emph{\textit{2016 IEEE Conference on Computer Vision and Pattern Recognition (CVPR)}}, pages 770--778, 2016.
\newblock \href{https://doi.org/10.1109/CVPR.2016.90}{\ttfamily\path{ doi:10.1109/CVPR.2016.90}}.

\bibitem[He et~al.(2017)He, Gkioxari, Dollár, and Girshick]{he2018mask}
Kaiming He, Georgia Gkioxari, Piotr Dollár, and Ross Girshick.
\newblock Mask r-cnn.
\newblock In \emph{\textit{2017 IEEE International Conference on Computer Vision (ICCV)}}, pages 2980--2988, 2017.
\newblock \href{https://doi.org/10.1109/ICCV.2017.322}{\ttfamily\path{ doi:10.1109/ICCV.2017.322}}.

\bibitem[Heusel et~al.(2017)Heusel, Ramsauer, Unterthiner, Nessler, and Hochreiter]{fid}
Martin Heusel, Hubert Ramsauer, Thomas Unterthiner, Bernhard Nessler, and Sepp Hochreiter.
\newblock Gans trained by a two time-scale update rule converge to a local nash equilibrium, 2017.

\bibitem[Hou et~al.(2019)Hou, Ma, Liu, and Loy]{hou2019learning}
Yuenan Hou, Zheng Ma, Chunxiao Liu, and Chen~Change Loy.
\newblock Learning lightweight lane detection cnns by self attention distillation.
\newblock In \emph{\textit{2019 IEEE/CVF International Conference on Computer Vision (ICCV)}}, pages 1013--1021, 2019.
\newblock \href{https://doi.org/10.1109/ICCV.2019.00110}{\ttfamily\path{ doi:10.1109/ICCV.2019.00110}}.

\bibitem[Isola et~al.(2017)Isola, Zhu, Zhou, and Efros]{pix2pix}
Phillip Isola, Jun-Yan Zhu, Tinghui Zhou, and Alexei~A Efros.
\newblock Image-to-image translation with conditional adversarial networks.
\newblock \emph{\textit{CVPR}}, 2017.

\bibitem[Jaunet et~al.(2021)Jaunet, Bono, Vuillemot, and Wolf]{jaunet2021sim2realvizvisualizingsim2realgap}
Theo Jaunet, Guillaume Bono, Romain Vuillemot, and Christian Wolf.
\newblock Sim2realviz: Visualizing the sim2real gap in robot ego-pose estimation, 2021.

\bibitem[Johnson et~al.(2021)Johnson, Douze, and Jégou]{faiss}
Jeff Johnson, Matthijs Douze, and Hervé Jégou.
\newblock Billion-scale similarity search with gpus.
\newblock \emph{\textit{IEEE Transactions on Big Data}}, 7\penalty0 (3):\penalty0 535--547, 2021.
\newblock \href{https://doi.org/10.1109/TBDATA.2019.2921572}{\ttfamily\path{ doi:10.1109/TBDATA.2019.2921572}}.

\bibitem[Kar et~al.(2019)Kar, Prakash, Liu, Cameracci, Yuan, Rusiniak, Acuna, Torralba, and Fidler]{metasim}
Amlan Kar, Aayush Prakash, Ming-Yu Liu, Eric Cameracci, Justin Yuan, Matt Rusiniak, David Acuna, Antonio Torralba, and Sanja Fidler.
\newblock Meta-sim: Learning to generate synthetic datasets.
\newblock In \emph{\textit{2019 IEEE/CVF International Conference on Computer Vision (ICCV)}}, pages 4550--4559, 2019.
\newblock \href{https://doi.org/10.1109/ICCV.2019.00465}{\ttfamily\path{ doi:10.1109/ICCV.2019.00465}}.

\bibitem[Kingma and Welling(2022)]{kingma2022autoencoding}
Diederik~P Kingma and Max Welling.
\newblock Auto-encoding variational bayes.
\newblock \href{https://arxiv.org/abs/1312.6114}{arXiv:1312.6114}, 2022.

\bibitem[Lambert et~al.(2020)Lambert, Liu, Sener, Hays, and Koltun]{mseg1}
John Lambert, Zhuang Liu, Ozan Sener, James Hays, and Vladlen Koltun.
\newblock {MSeg}: A composite dataset for multi-domain semantic segmentation.
\newblock In \emph{\textit{Computer Vision and Pattern Recognition (CVPR)}}, 2020.

\bibitem[Lillicrap et~al.(2019)Lillicrap, Hunt, Pritzel, Heess, Erez, Tassa, Silver, and Wierstra]{lillicrap2019continuous}
Timothy~P. Lillicrap, Jonathan~J. Hunt, Alexander Pritzel, Nicolas Heess, Tom Erez, Yuval Tassa, David Silver, and Daan Wierstra.
\newblock Continuous control with deep reinforcement learning.
\newblock \href{https://arxiv.org/abs/1509.02971}{arXiv:1509.02971}, 2019.

\bibitem[Lin et~al.(2014)Lin, Maire, Belongie, Bourdev, Girshick, Hays, Perona, Ramanan, Doll{'{a} }r, and Zitnick]{cocodataset}
Tsung{-}Yi Lin, Michael Maire, Serge~J. Belongie, Lubomir~D. Bourdev, Ross~B. Girshick, James Hays, Pietro Perona, Deva Ramanan, Piotr Doll{'{a} }r, and C.~Lawrence Zitnick.
\newblock Microsoft {COCO:} common objects in context.
\newblock \emph{\textit{CoRR}}, abs/1405.0312, 2014.
\newblock \href{http://arxiv.org/abs/1405.0312}{{\ttfamily arXiv:1405.0312}}.

\bibitem[Liu et~al.(2018)Liu, Breuel, and Kautz]{liu2018unsupervisedimagetoimagetranslationnetworks}
Ming-Yu Liu, Thomas Breuel, and Jan Kautz.
\newblock Unsupervised image-to-image translation networks, 2018.

\bibitem[Mahajan et~al.(2024)Mahajan, Unjhawala, Zhang, Zhou, Young, Ruiz, Caldararu, Batagoda, Ashokkumar, and Negrut]{mahajan2024quantifyingsim2realgapgps}
Ishaan Mahajan, Huzaifa Unjhawala, Harry Zhang, Zhenhao Zhou, Aaron Young, Alexis Ruiz, Stefan Caldararu, Nevindu Batagoda, Sriram Ashokkumar, and Dan Negrut.
\newblock Quantifying the sim2real gap for gps and imu sensors, 2024.

\bibitem[Mittermueller et~al.(2022)Mittermueller, Ye, and Hlavacs]{9893673}
Martina Mittermueller, Zhanxiang Ye, and Helmut Hlavacs.
\newblock Est-gan: Enhancing style transfer gans with intermediate game render passes.
\newblock In \emph{2022 IEEE Conference on Games (CoG)}, pages 25--32, 2022.
\newblock \href{https://doi.org/10.1109/CoG51982.2022.9893673}{\ttfamily\path{ doi:10.1109/CoG51982.2022.9893673}}.

\bibitem[Mnih et~al.(2013)Mnih, Kavukcuoglu, Silver, Graves, Antonoglou, Wierstra, and Riedmiller]{mnih2013playing}
Volodymyr Mnih, Koray Kavukcuoglu, David Silver, Alex Graves, Ioannis Antonoglou, Daan Wierstra, and Martin Riedmiller.
\newblock Playing atari with deep reinforcement learning.
\newblock \href{https://arxiv.org/abs/1312.5602}{arXiv:1312.5602}, 2013.

\bibitem[Neuhold et~al.(2017)Neuhold, Ollmann, Bulò, and Kontschieder]{vistas}
Gerhard Neuhold, Tobias Ollmann, Samuel~Rota Bulò, and Peter Kontschieder.
\newblock The mapillary vistas dataset for semantic understanding of street scenes.
\newblock In \emph{\textit{2017 IEEE International Conference on Computer Vision (ICCV)}}, pages 5000--5009, 2017.
\newblock \href{https://doi.org/10.1109/ICCV.2017.534}{\ttfamily\path{ doi:10.1109/ICCV.2017.534}}.

\bibitem[Pahk et~al.(2023)Pahk, Shim, Baek, Lim, and Choi]{10086509}
Jinu Pahk, Jungseok Shim, Minhyeok Baek, Yongseob Lim, and Gyeungho Choi.
\newblock Effects of sim2real image translation via dclgan on lane keeping assist system in carla simulator.
\newblock \emph{\textit{IEEE Access}}, 11:\penalty0 33915--33927, 2023.
\newblock \href{https://doi.org/10.1109/ACCESS.2023.3262991}{\ttfamily\path{ doi:10.1109/ACCESS.2023.3262991}}.

\bibitem[Park et~al.(2019)Park, Liu, Wang, and Zhu]{park2019semantic}
Taesung Park, Ming-Yu Liu, Ting-Chun Wang, and Jun-Yan Zhu.
\newblock Semantic image synthesis with spatially-adaptive normalization.
\newblock In \emph{\textit{2019 IEEE/CVF Conference on Computer Vision and Pattern Recognition (CVPR)}}, pages 2332--2341, 2019.
\newblock \href{https://doi.org/10.1109/CVPR.2019.00244}{\ttfamily\path{ doi:10.1109/CVPR.2019.00244}}.

\bibitem[Park et~al.(2020)Park, Efros, Zhang, and Zhu]{park2020contrastive}
Taesung Park, Alexei~A. Efros, Richard Zhang, and Jun-Yan Zhu.
\newblock Contrastive learning for unpaired image-to-image translation.
\newblock In Andrea Vedaldi, Horst Bischof, Thomas Brox, and Jan-Michael Frahm, editors, \emph{\textit{Computer Vision -- ECCV 2020}}, pages 319--345, Cham, 2020. Springer International Publishing.

\bibitem[Paszke et~al.(2019)Paszke, Gross, Massa, Lerer, Bradbury, Chanan, Killeen, Lin, Gimelshein, Antiga, Desmaison, Kopf, Yang, DeVito, Raison, Tejani, Chilamkurthy, Steiner, Fang, Bai, and Chintala]{NEURIPS2019_9015}
Adam Paszke, Sam Gross, Francisco Massa, Adam Lerer, James Bradbury, Gregory Chanan, Trevor Killeen, Zeming Lin, Natalia Gimelshein, Luca Antiga, Alban Desmaison, Andreas Kopf, Edward Yang, Zachary DeVito, Martin Raison, Alykhan Tejani, Sasank Chilamkurthy, Benoit Steiner, Lu~Fang, Junjie Bai, and Soumith Chintala.
\newblock Pytorch: An imperative style, high-performance deep learning library.
\newblock In \emph{\textit{Advances in Neural Information Processing Systems 32}}, pages 8024--8035. Curran Associates, Inc., 2019.

\bibitem[Ram et~al.(2022)Ram, Bakker, and Lew]{ddpg-sim2real}
Jochem Ram, E.M. Bakker, and Michael~S. Lew.
\newblock Sim-to-real autonomous driving in carla using image translation and deep deterministic policy gradient.
\newblock \url{https://theses.liacs.nl/pdf/2021-2022-RamJ.pdf}, 2022.

\bibitem[Ranftl et~al.(2022)Ranftl, Lasinger, Hafner, Schindler, and Koltun]{ranftl2020robust}
R.~Ranftl, K.~Lasinger, D.~Hafner, K.~Schindler, and V.~Koltun.
\newblock Towards robust monocular depth estimation: Mixing datasets for zero-shot cross-dataset transfer.
\newblock \emph{\textit{IEEE Transactions on Pattern Analysis \& Machine Intelligence}}, 44\penalty0 (03):\penalty0 1623--1637, mar 2022.
\newblock ISSN 1939-3539.
\newblock \href{https://doi.org/10.1109/TPAMI.2020.3019967}{\ttfamily\path{ doi:10.1109/TPAMI.2020.3019967}}.

\bibitem[Redmon et~al.(2016)Redmon, Divvala, Girshick, and Farhadi]{yolo}
Joseph Redmon, Santosh Divvala, Ross Girshick, and Ali Farhadi.
\newblock You only look once: Unified, real-time object detection.
\newblock In \emph{\textit{2016 IEEE Conference on Computer Vision and Pattern Recognition (CVPR)}}, pages 779--788, 2016.
\newblock \href{https://doi.org/10.1109/CVPR.2016.91}{\ttfamily\path{ doi:10.1109/CVPR.2016.91}}.

\bibitem[Richter et~al.(2016)Richter, Vineet, Roth, and Koltun]{Richter_2016_ECCV}
Stephan~R. Richter, Vibhav Vineet, Stefan Roth, and Vladlen Koltun.
\newblock Playing for data: {G}round truth from computer games.
\newblock In Bastian Leibe, Jiri Matas, Nicu Sebe, and Max Welling, editors, \emph{\textit{European Conference on Computer Vision (ECCV)}}, volume 9906 of \emph{LNCS}, pages 102--118. Springer International Publishing, 2016.

\bibitem[Richter et~al.(2023)Richter, Alhaija, and Koltun]{Richter_2021}
Stephan~R. Richter, Hassan~Abu Alhaija, and Vladlen Koltun.
\newblock Enhancing photorealism enhancement.
\newblock \emph{IEEE Transactions on Pattern Analysis and Machine Intelligence}, 45\penalty0 (2):\penalty0 1700--1715, 2023.
\newblock \href{https://doi.org/10.1109/TPAMI.2022.3166687}{\ttfamily\path{ doi:10.1109/TPAMI.2022.3166687}}.

\bibitem[Ronneberger et~al.(2015)Ronneberger, Fischer, and Brox]{ronneberger2015unet}
Olaf Ronneberger, Philipp Fischer, and Thomas Brox.
\newblock U-net: Convolutional networks for biomedical image segmentation.
\newblock In Nassir Navab, Joachim Hornegger, William~M. Wells, and Alejandro~F. Frangi, editors, \emph{\textit{Medical Image Computing and Computer-Assisted Intervention -- MICCAI 2015}}, pages 234--241, Cham, 2015. Springer International Publishing.

\bibitem[Ros et~al.(2016)Ros, Sellart, Materzynska, Vazquez, and Lopez]{7780721}
German Ros, Laura Sellart, Joanna Materzynska, David Vazquez, and Antonio~M. Lopez.
\newblock The synthia dataset: A large collection of synthetic images for semantic segmentation of urban scenes.
\newblock In \emph{2016 IEEE Conference on Computer Vision and Pattern Recognition (CVPR)}, pages 3234--3243, 2016.
\newblock \href{https://doi.org/10.1109/CVPR.2016.352}{\ttfamily\path{ doi:10.1109/CVPR.2016.352}}.

\bibitem[Selvaraju et~al.(2017)Selvaraju, Cogswell, Das, Vedantam, Parikh, and Batra]{gradcam}
Ramprasaath~R. Selvaraju, Michael Cogswell, Abhishek Das, Ramakrishna Vedantam, Devi Parikh, and Dhruv Batra.
\newblock Grad-cam: Visual explanations from deep networks via gradient-based localization.
\newblock In \emph{2017 IEEE International Conference on Computer Vision (ICCV)}, pages 618--626, 2017.
\newblock \href{https://doi.org/10.1109/ICCV.2017.74}{\ttfamily\path{ doi:10.1109/ICCV.2017.74}}.

\bibitem[Shao et~al.(2022)Shao, Wang, Chen, Li, and Liu]{shao2022interfuser}
Hao Shao, Letian Wang, RuoBing Chen, Hongsheng Li, and Yu~Liu.
\newblock Safety-enhanced autonomous driving using interpretable sensor fusion transformer.
\newblock \href{https://arxiv.org/abs/2207.14024}{arXiv:2207.14024}, 2022.

\bibitem[Shao et~al.(2023)Shao, Wang, Chen, Waslander, Li, and Liu]{shao2023reasonnet}
Hao Shao, Letian Wang, Ruobing Chen, Steven~L Waslander, Hongsheng Li, and Yu~Liu.
\newblock Reasonnet: End-to-end driving with temporal and global reasoning.
\newblock In \emph{\textit{Proceedings of the IEEE/CVF Conference on Computer Vision and Pattern Recognition}}, pages 13723--13733, 2023.

\bibitem[Simonyan and Zisserman(2015)]{simonyan2015deep}
Karen Simonyan and Andrew Zisserman.
\newblock Very deep convolutional networks for large-scale image recognition.
\newblock In \emph{\textit{International Conference on Learning Representations}}, 2015.

\bibitem[Sushko et~al.(2022)Sushko, Sch{\"o}nfeld, Zhang, Gall, Schiele, and Khoreva]{Sushko2022}
Vadim Sushko, Edgar Sch{\"o}nfeld, Dan Zhang, Juergen Gall, Bernt Schiele, and Anna Khoreva.
\newblock Oasis: Only adversarial supervision for semantic image synthesis.
\newblock \emph{International Journal of Computer Vision}, 130\penalty0 (12):\penalty0 2903--2923, 2022.
\newblock ISSN 1573-1405.
\newblock \href{https://doi.org/10.1007/s11263-022-01673-x}{\ttfamily\path{ doi:10.1007/s11263-022-01673-x}}.

\bibitem[Tang et~al.(2023)Tang, Liu, Xu, Torr, and Sebe]{tang2021attentiongan}
Hao Tang, Hong Liu, Dan Xu, Philip H.~S. Torr, and Nicu Sebe.
\newblock Attentiongan: Unpaired image-to-image translation using attention-guided generative adversarial networks.
\newblock \emph{\textit{IEEE Transactions on Neural Networks and Learning Systems}}, 34\penalty0 (4):\penalty0 1972--1987, 2023.
\newblock \href{https://doi.org/10.1109/TNNLS.2021.3105725}{\ttfamily\path{ doi:10.1109/TNNLS.2021.3105725}}.

\bibitem[Theiss et~al.(2022)Theiss, Leverett, Kim, and Prakash]{zheng2022vsait}
Justin~D. Theiss, Jay Leverett, Daeil Kim, and Aayush Prakash.
\newblock Unpaired image translation via vector symbolic architectures.
\newblock In \emph{European Conference on Computer Vision}, 2022.

\bibitem[Xiong et~al.(2023)Xiong, Xiao, Yao, Liu, Yang, and Fu]{10210278}
Yonggang Xiong, Xueming Xiao, Meibao Yao, Haiqiang Liu, Hong Yang, and Yuegang Fu.
\newblock Marsformer: Martian rock semantic segmentation with transformer.
\newblock \emph{IEEE Transactions on Geoscience and Remote Sensing}, 61:\penalty0 1--12, 2023.
\newblock \href{https://doi.org/10.1109/TGRS.2023.3302649}{\ttfamily\path{ doi:10.1109/TGRS.2023.3302649}}.

\bibitem[Xiong et~al.(2024)Xiong, Xiao, Yao, Cui, and Fu]{XIONG2024167}
Yonggang Xiong, Xueming Xiao, Meibao Yao, Hutao Cui, and Yuegang Fu.
\newblock Light4mars: A lightweight transformer model for semantic segmentation on unstructured environment like mars.
\newblock \emph{ISPRS Journal of Photogrammetry and Remote Sensing}, 214:\penalty0 167--178, 2024.
\newblock ISSN 0924-2716.
\newblock \href{https://doi.org/https://doi.org/10.1016/j.isprsjprs.2024.06.008}{\ttfamily\path{ doi:https://doi.org/10.1016/j.isprsjprs.2024.06.008}}.

\bibitem[Yeh et~al.(2022)Yeh, Nagano, Khamis, Kautz, Liu, and Wang]{3555442}
Yu-Ying Yeh, Koki Nagano, Sameh Khamis, Jan Kautz, Ming-Yu Liu, and Ting-Chun Wang.
\newblock Learning to relight portrait images via a virtual light stage and synthetic-to-real adaptation.
\newblock \emph{ACM Trans. Graph.}, 41\penalty0 (6), November 2022.
\newblock ISSN 0730-0301.
\newblock \href{https://doi.org/10.1145/3550454.3555442}{\ttfamily\path{ doi:10.1145/3550454.3555442}}.
\newblock URL \url{https://doi.org/10.1145/3550454.3555442}.

\bibitem[Yu et~al.(2024)Yu, Foote, Mooney, and Martín-Martín]{yu2024naturallanguagehelpbridge}
Albert Yu, Adeline Foote, Raymond Mooney, and Roberto Martín-Martín.
\newblock Natural language can help bridge the sim2real gap, 2024.

\bibitem[Zhang et~al.(2023)Zhang, Rao, and Agrawala]{zhang2023adding}
Lvmin Zhang, Anyi Rao, and Maneesh Agrawala.
\newblock Adding conditional control to text-to-image diffusion models., 2023.

\bibitem[Zhao et~al.(2025)Zhao, Yao, Zhao, Jiang, Zhang, Xiao, and Gao]{m2cs}
Huanfeng Zhao, Meibao Yao, Yan Zhao, Yao Jiang, Hongyan Zhang, Xueming Xiao, and Ke~Gao.
\newblock M2cs: A multimodal and campus-scapes dataset for dynamic slam and moving object perception.
\newblock \emph{Journal of Field Robotics}, 42\penalty0 (3):\penalty0 787--805, 2025.

\bibitem[Zhu et~al.(2017)Zhu, Park, Isola, and Efros]{zhu2020unpaired}
Jun-Yan Zhu, Taesung Park, Phillip Isola, and Alexei~A. Efros.
\newblock Unpaired image-to-image translation using cycle-consistent adversarial networks.
\newblock In \emph{\textit{2017 IEEE International Conference on Computer Vision (ICCV)}}, pages 2242--2251, 2017.
\newblock \href{https://doi.org/10.1109/ICCV.2017.244}{\ttfamily\path{ doi:10.1109/ICCV.2017.244}}.

\end{thebibliography}
	
\end{document}